%% file: 00-compgen.tex
\pdfoutput=1

\documentclass[11pt]{article}

\usepackage[preprint]{acl}

\usepackage{times}
\usepackage{latexsym}
\usepackage[T1]{fontenc}
\usepackage[utf8]{inputenc}

\usepackage{microtype}

\usepackage{inconsolata}

\input{00-header.tex}

\title{\textsc{CoGen}: Learning from Feedback \\ with Coupled Comprehension and Generation}

\author{Mustafa Omer Gul \and Yoav  Artzi\\
  Department of Computer Science and Cornell Tech, Cornell University  \\  \texttt{\{momergul, yoav\}@cs.cornell.edu}}

\begin{document}
\maketitle

\begin{abstract}

\input{01-abstract.tex}

\end{abstract}

\input{10-intro}

\input{20-technical-overview}

\input{30-continual-learning}

\input{40-merging-compgen}

\input{50-experimental-setup}

\input{60-results}

\input{70-related-work}

\input{80-discussion}

\input{90-limitations-and-ethics}

\bibliography{custom}

\input{90-appendix}

\end{document}

%% file: 00-header.tex
\usepackage{graphicx}
\usepackage{xspace}

\usepackage{caption}
\usepackage{subcaption}

\newcommand{\aautoref}[1]{\hyperref[#1]{Appendix~\ref*{#1}}}
\usepackage{graphicx}

\usepackage{array}

\usepackage{tabularx}

\usepackage{tikz}

\usepackage[inkscapearea=page]{svg}

\usepackage{adjustbox}

\usepackage{float}

\usepackage{booktabs}

\usepackage{amsmath}
\usepackage{amsfonts}

\usepackage{arydshln}
\usepackage{bbm}
\usepackage{hyperref}
\usepackage{pgfplots}
\usepackage{algorithm}
\usepackage[noend]{algpseudocode}
\usepackage{enumitem}
\usepackage{array}
\usepgfplotslibrary{fillbetween}
\usepgfplotslibrary{groupplots}
\pgfplotsset{compat=1.3}
\usetikzlibrary{calc}
\usetikzlibrary{fit}
\usetikzlibrary{matrix}
\usepackage{multicol}
 \usepackage{vwcol}  
\usepackage{float}

\newcolumntype{S}{>{\hsize=.5\hsize}X}
\newcolumntype{M}{>{\hsize=.333\hsize}X}
\newcolumntype{L}{>{\hsize=2.5\hsize}X}

\newcolumntype{U}{>{\hsize=\hsize}X}
\newcolumntype{D}{>{\hsize=2\hsize}X}

\newcommand{\system}[1]{\textsc{#1}\xspace}
\newcommand{\variantfull}{\system{Full}}
\newcommand{\variantnods}{\system{No-DS}}
\newcommand{\variantnoji}{\system{No-JI}}
\newcommand{\variantbaseline}{\system{Baseline}}
\newcommand{\varianthuman}{\system{Human}}

\usepackage{titlesec}
\titlespacing*{\paragraph} {0pt}{.8ex plus .1ex minus .1ex}{1em}

\newcommand{\eat}[1]{}

\newcommand{\nlstring}[1]{\textit{#1}}

\newcommand{\kilogram}{\textsc{KiloGram}\xspace}

\newcommand{\imageset}{\mathcal{I}}
\newcommand{\image}{I}
\newcommand{\utterance}{u}

\newcommand{\reward}{r}
\newcommand{\targetindex}{t}

\newcommand{\speaker}{s}
\newcommand{\listener}{l}
\newcommand{\joint}{j}

\newcommand{\dataset}{\mathcal{D}}

\newcommand{\round}{\rho}

\newcommand{\idefics}{IDEFICS2\xspace}

\newcommand{\params}{\theta}

%% file: 01-abstract.tex
Systems with both language comprehension and generation capabilities can benefit from the tight connection between the two. This work studies coupling comprehension and generation with focus on continually learning from interaction with users. We propose techniques to tightly integrate the two capabilities for both learning and inference. We situate our studies in two-player reference games, and deploy various models for thousands of interactions with human users, while learning from interaction feedback signals. We show dramatic improvements in performance over time, with comprehension-generation coupling leading to performance improvements up to 26\% in absolute terms and up to 17\% higher accuracies compared to a non-coupled system. Our analysis also shows coupling has substantial qualitative impact on the system's language, making it significantly more human-like.

%% file: 10-intro.tex
\section{Introduction}\label{sec:intro}
Language comprehension and generation are closely related processes. Indeed, observations such as the ability to finish incomplete partner utterances in dialogue~\cite{clark1986referring, howes2011incrementality}, as well as neuroscientific evidence~\citep{paus1996modulation, opitz2003phonological, menenti2011shared} have led to integrated accounts of comprehension and generation in cognitive science~\cite{pickering2013integrated,pickering2018predicting}, where processes related to generation are active during comprehension and vice versa. 
This suggests a potential for coupling the two in computational systems, and creating a virtuous cycle, where the improvement of one capability drives learning and performance in the other. 
This is particularly compelling in systems that continually learn and improve through interaction with users, where the dynamics between the two capabilities play out over time.

We study the dynamics of this coupling in a continual learning\footnote{\emph{Continual learning} is at times used to describe scenarios where models are adapted to new tasks. We use it in the sense of improving a model on its original task over time.} setting, where trends in learning and behavior can be observed over time. We design an interaction scenario where models can take both listener (comprehension) and speaker (generation) roles, and receive feedback while interacting with human partners. We couple comprehension and generation through several mechanisms, and observe the impact this coupling has on the long-term dynamics of performance and language. 

\begin{figure}
    \centering
    \includegraphics[width=0.7\columnwidth,clip,trim=188 233 378 110]{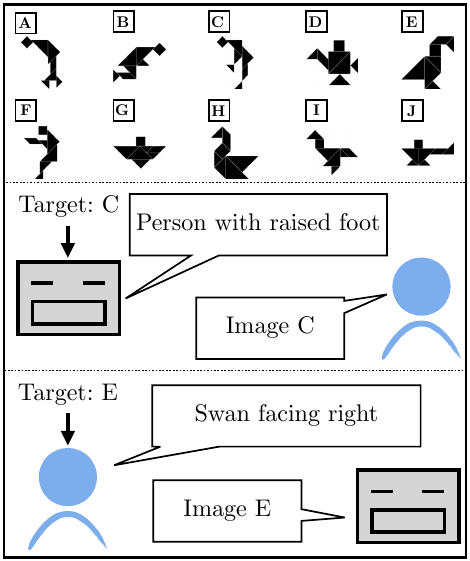}
    \caption{Illustration of our reference game interaction scenario involving a speaker and listener. Each game includes a single turn.   Speakers are assigned a target image and write a description such that their partner can guess the image from the description. The game succeeds if the listener guesses correctly. We deploy our models (gray bot) as speaker to interact with human listeners (top) or vice versa (bottom).}
    \label{fig:interaction}
    \vspace{-5pt}
\end{figure}

\begin{figure*}
    \centering
    \includegraphics[width=0.9\textwidth,clip,trim=32 225 60 120]{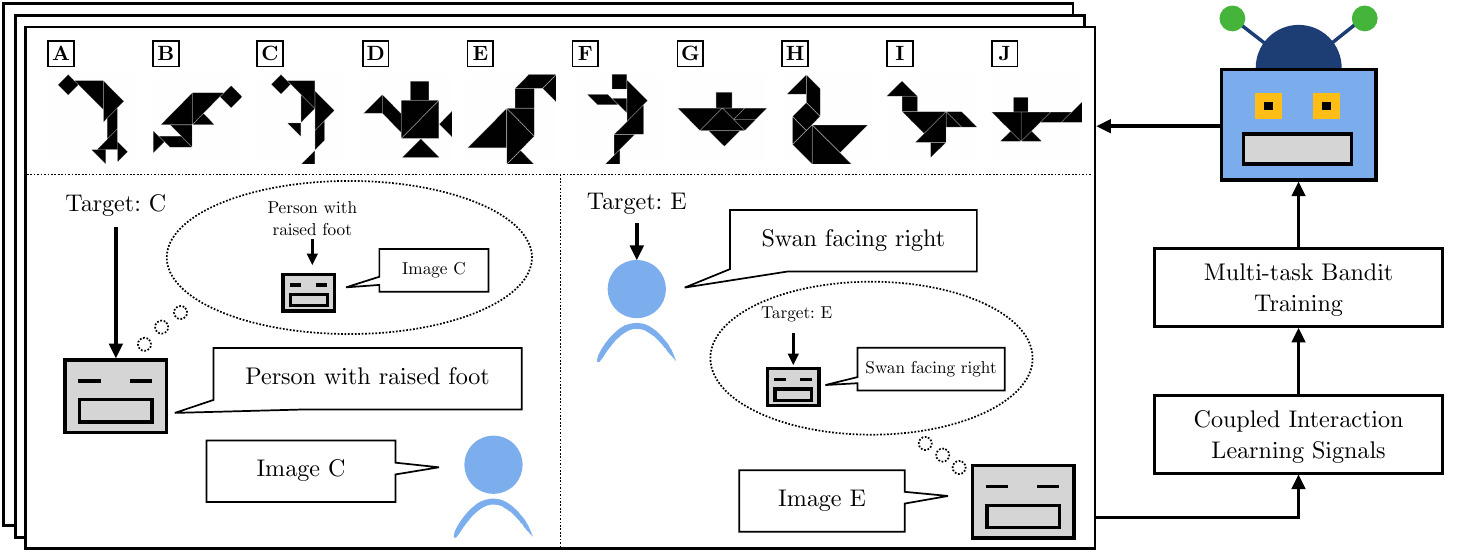}
    \caption{Illustration of our continual learning scenario with coupled comprehension and generation. The process alternates between interactions with human partners in a reference game, and training using learning signals from the interactions. The model performs both the generation (left) and comprehension (right) tasks, while jointly reasoning over the other role (thought bubbles). Training leverages feedback for the role the model performs as well as the opposing role. Following each round of training, we re-deploy the updated model and repeat the process.}
    \label{fig:intro}
    \vspace{-5pt}
\end{figure*}

We instantiate comprehension and generation as the listener and speaker roles of a two-player reference game~\citep{krauss1964changes, clark1986referring} involving abstract visual stimuli~\citep{ji2022abstract}, which remain challenging for state-of-the-art vision-language models (\autoref{fig:interaction}). 
We deploy a single model that can take both roles. 
The process alternates between the model interacting with human partners, and training to improve both comprehension and generation capabilities based on feedback from the interactions.

We couple comprehension and generation through two strategies: (a) at inference-time via a joint inference process that incorporates the opposing role, and (b) at training time by generating examples and rewards for each role from feedback on performance in that role as well as the opposing role. 
\autoref{fig:intro} illustrates the deployment and coupling mechanisms. 
The combination of these strategies creates a virtuous cycle that evolves over time, as the system continually trains and improves. 
As one capability improves (e.g., comprehension), the model's performance on the opposing capability (e.g., generation) also improves via the joint inference procedure. This, in turn, leads to better interactions, and the feedback the model receives changes as its capabilities advance and its failure modes change. 
The coupling of feedback signals via the training data qualitatively changes the training beyond a simple increase in the amount of data. Whereas a generation system that trains on feedback is only exposed to its own language, the coupled system is continually exposed to a stream of new human language. This can enable the system to expand its generation abilities and make the language more similar to humans, beyond simply refining it to maximize interaction performance. 

We conduct extensive experiments, concurrently deploying a baseline and multiple model variants for thousands of interactions with human partners in a controlled study. 
Our focus is to observe both performance and language trends over time. 
Our coupled approach shows dramatic and fast performance gains, overall improving by 19.48\% for comprehension and 26.07\% for generation, in absolute terms. 
At the conclusion of our deployment, the coupled approach outperforms the non-coupled baseline by 14.80\% for comprehension and 17.10\% for generation. Furthermore, coupling results in greater data efficiency, with the full system still outperforming this baseline with less than one-third the number of human interactions.
We observe coupling dramatically influences the generated language, with the coupled approach exhibiting a larger effective vocabulary and greater alignment with human language according to both linguistic measures such as utterance length and automated metrics such as MAUVE~\citep{pillutla2021mauve}.
Our code, data, and experiment logs are available at \url{https://github.com/lil-lab/cogen}.

%% file: 20-technical-overview.tex
\section{Interaction Scenario and Overview}\label{sec:overview_and_scenario}

We study the coupling of comprehension and generation by training and deploying an agent that interacts with human users, and continually learns from these interactions. 
This allows us to observe how the interplay between comprehension and generation evolves over time, and what the long-term effects of coupling the two processes are.

\paragraph{Interaction Scenario}

We use a reference game as our interaction scenario (\autoref{fig:interaction}).
Each game involves two players: a speaker and listener. 
Both participants are presented a set of abstract tangram images as context $\imageset = \{\image_1, \ldots, \image_N\}$. 
Each participant observes the images in a different order. 
The speaker is given a target $\image_\targetindex \in \imageset$, and generates an utterance $\utterance$, with the goal of allowing the listener to pick the target $\image_\targetindex$ from the set of the images. The listener then makes a choice. An interaction succeeds if the listener picks the intended target. 

Reference games have been extensively used in research, including in NLP~\cite[e.g.,][]{andreas-klein-2016-reasoning, ji2022abstract} and cognitive science~\cite[e.g.,][]{krauss1964changes, rosenberg1964speakers, clark1986referring, hawkins2023partners}, and provide a balance between complexity and research feasibility: (a) the interaction includes both generation and comprehension; (b) they are relatively accessible for crowdsourcing workers; (c) they are well scoped so learning is feasible without excessive data requirements; and (d) success is easy to measure. 
The tangram shapes we use have been shown to elicit rich linguistic behavior, both from listeners and speakers~\citep{schober1989understanding, ji2022abstract}. 
They also remain challenging for contemporary models. Our models' initial performance is at least 33.1\% below human accuracy (\autoref{sec:results}).

\paragraph{Deployment}

We deploy our model to interact with humans in rounds. 
Each round includes a predetermined number of interactions, each with the model taking one of the roles (speaker or listener) and the human participant taking the other. 
We derive feedback signals from the interactions, construct training examples, and train our model.
Following training, we re-deploy for the next round.

\paragraph{Inference and Learning} 

We use \idefics-8B~\citep{laurenccon2024matters} as our model, an auto-regressive LLM that can also take images as part of its input. 
The model is parameterized by $\params$. 
As a speaker (generation), the model computes a probability distribution $P_{\speaker}(\utterance \vert \imageset, \targetindex; \params)$ over descriptions $\utterance$  of the target $\targetindex \in \{1,\dots,N\}$ in the given context $\imageset$. 
As a listener (comprehension), it computes a distribution $P_{\listener}(\targetindex\vert\imageset,\utterance; \params)$ over target $\targetindex$ selections, given the context $\imageset$ and a description utterance $\utterance$.  
Both utterances and target selections are generated via a conventional auto-regressive process. 
Following each deployment round, we train the model using all the feedback data collected so far by treating the feedback as rewards for contextual bandit learning. 

\paragraph{Evaluation} 

Our main evaluation is conducted through interaction with humans, where each round forms the evaluation of the model so far. We evaluate the comprehension performance of the model from its target selection accuracy as a listener. We evaluate generation as the accuracy of the human listener in selecting the target given the model's generated description when in the speaker role. 
We also study the linguistic trends of the model's generations over time to better understand the dynamics created by coupling comprehension and generation. 
This includes analyzing its similarity to human language and its linguistic properties.

%% file: 30-continual-learning.tex
\section{Continual Learning}\label{sec:learning}

We combine the continual learning approaches of \citet{kojima2021continual} (generation) and \citet{suhr2024continual} (comprehension).
Both approaches map feedback to rewards, treat learning as a contextual bandit problem, and use REINFORCE~\cite{williams1992simple}, a relatively simple policy gradient algorithm. We adopt these design choices. 
A key difference of our process is that we combine the comprehension and generation objectives to train a single model. 

Deployment and learning are interleaved. Each round $\round$ starts with deploying the model parameterized by $\params_\round$ to collect interactions with humans. We record feedback signals from these interactions. Upon collecting a set of interactions, we re-train the model given all data collected so far to estimate new parameters $\params_{\round+1}$. The model is then deployed for the next round, and the process continues.

\subsection{Feedback Collection}\label{sec:learn:feedback}

Feedback collection is part of the model interacting with human partners (i.e., the system deployment), and differs depending on the model's role. 
As the listener, the model is given context $\imageset$ and a human-generated utterance $\utterance$ and predicts the index of the target image $\hat{\targetindex} = \arg\max_\targetindex P_\listener(\targetindex \vert \imageset, \utterance; \params_\round)$. 
The game then indicates if the selection was correct or not, and terminates. 
We treat this indication as feedback,\footnote{In single-turn reference games, such as in our scenario, task success and feedback are the same, so we do not solicit explicit feedback. However, our approach is designed to be applicable to settings where feedback and task success do not collapse to be the same, such as the setting considered by \citet{kojima2021continual} and \citet{suhr2024continual}.} and directly map it to a binary reward to create a comprehension datapoint: $(\imageset, \utterance, \hat{\targetindex}, \reward)$, with $\reward=1$ upon game success and $\reward=-1$ otherwise. 
Likewise, as the speaker, the model samples an utterance $\hat{\utterance} \sim P_\speaker(\utterance\vert\imageset, \targetindex; \params_\round)$ given context $\imageset$ and target image index $\targetindex$. 
The game indication of success provides the feedback, resulting in a generation datapoint: $(\imageset, \hat{\utterance}, \targetindex, \reward)$ with $\reward \in \{-1,1\}$ accordingly. 
Each round results  in two datasets: $\dataset_{\listener, \round}$ and $\dataset_{\speaker, \round}$ for comprehension and generation.
In both, datapoints constitute model output produced during interaction, and a reward for it. 
This is in contrast to supervised learning, where datapoints include output annotations, or human feedback as used in RLHF, where datapoints are pairwise preferences drawn from external annotators. 

\subsection{Learning}\label{sec:parameter_optimization}

We estimate the next round's model parameters $\theta_{\rho+1}$ by re-training from the initial weights (i.e., the original IDEFICS2 weights).
The comprehension training dataset is a union of all collected feedback data so far $\dataset_{\listener, \leq \round} =  \bigcup^\round_{i=1}D_{\listener,i}$. 
The production task dataset $\dataset_{\speaker, \leq \round}$ is similarly defined.

We frame learning as a contextual bandit problem with a multi-task additive objective combining the comprehension and generation components. We optimize with a REINFORCE-style policy gradient algorithm~\citep{williams1992simple}. This choice follows prior work~\citep{kojima2021continual, suhr2024continual}, and is motivated by the simplicity of REINFORCE, critical in a setting where humans are part of the iterative learning process.\footnote{More generally, \citet{ahmadian2024basics} recently showed REINFORCE can match more modern methods, such as PPO.} 
The gradient for a comprehension example $(\imageset, \utterance, \hat{\targetindex}, \reward) \sim \dataset_{\listener, \leq \round}$ collected at round $m$ is:
\begin{equation}
    \Delta_\listener = c_\listener \reward \nabla \log P_\listener(\hat{\targetindex} \vert \imageset, \utterance ; \params)\;\;,
\end{equation}
where $c_\listener$ is the cased inverse propensity score (IPS) coefficient introduced by \citet{kojima2021continual} to mitigate the effect of negative examples (i.e., $\reward = -1$) allowing for unbounded loss:
\begin{equation}
    c_\listener = \begin{cases}
        \frac{P_\listener(\hat{\targetindex} \vert \imageset, \utterance; \params)}{P_\listener(\hat{\targetindex} \vert \imageset, \utterance; \params_m)} & \text{if } \reward = -1 \\
        1 & \text{else}
    \end{cases}\;\;,
\end{equation}
where $P_\listener(\hat{\targetindex} \vert \imageset, \utterance; \params_m)$ is the probability of the target $\hat{\targetindex}$ when it was sampled during the interaction at round $m$.
Without this coefficient, negative examples (i.e., $\reward = -1$) can dominate the loss and destabilize learning as their probabilities decrease, because $\lim_{P_\listener(\cdot) \to 0}\log P_\listener(\cdot) = -\infty$. 
The coefficient $c_\listener$ decreases the importance of such examples as their probability decreases. 
The gradient $\Delta_\speaker$ for generation datapoints $(\imageset, \hat{\utterance}, \targetindex, \reward)$ is identical, except using the generation distribution $P_\speaker(\hat{\utterance}\vert\imageset, \targetindex; \params)$.

%% file: 40-merging-compgen.tex
\section{Coupling Comprehension and Generation}\label{sec:couple}

We couple comprehension and production during both learning and inference. 
We also use one model for both tasks, creating a coupling at the parameter level, which is common in contemporary methods, partially due to high memory needs.

\subsection{Learning with Data Sharing}\label{sec:couple:datashare}

We convert comprehension datapoints to generation datapoints, and vice versa, to fully utilize the data models are exposed to in interactions. 
For example, consider the case of an agent in the role of a listener. 
If the speaker partner generates the utterance \nlstring{the target is a swan facing right} and the listener correctly guesses the target image (as in \autoref{fig:intro}), the listener does not only receive positive feedback for their guess, but also can learn that \nlstring{a swan facing right} is a valid description for the current context-target pair. 

Given datasets for comprehension $\dataset_{\listener, \round}$ and generation $\dataset_{\speaker, \round}$ collected at round $\round$, we expand both:
\begin{align}
    \dataset_{\listener, \round} &= \dataset_{\listener, \round} \cup \{(\imageset, \hat{\utterance}, \targetindex, \reward) \in \dataset_{\speaker, \round} \mid \reward = 1\}, \\
    \dataset_{\speaker, \round} &= \dataset_{\speaker, \round} \cup \{(\imageset, \utterance, \hat{\targetindex}, \reward) \in \dataset_{\listener, \round}  \mid \reward = 1\}.
\end{align}

We only convert positively labeled feedback ($\reward=1$), because we generally find positive rewards to be more reliable. 
A negative reward for a generated utterance could be because the utterance is incorrect or ambiguous, or the human listener made a mistake. 
The listener task is essentially classification. Creating a comprehension example with negative reward from such an example indicates to the model the utterance is a valid description for another target. This is a misleading signal, and in early pilot studies we found it not to be helpful, so we only convert examples with positive reward. 

An important result of this process is introducing human language into the training data of the speaker model. 
Generally, if a generating model learns from feedback only~\cite{kojima2021continual}, it is only exposed to language it has generated. This can lead to its language drifting from human language, even if its accuracy and legibility to human partners increase. 
Taking advantage of human utterances for the purpose of generation training opens up this closed system. We further discuss this in our results and analysis (\autoref{sec:results}).

\subsection{Joint Inference}\label{sec:couple:jointinference}

We couple the two distributions $P_\listener$ and $P_\speaker$ during inference by sampling from one distribution (i.e., $P_\listener$ in the case of comprehension) and then re-rank with a weighted geometric mean of the two distributions.
The weight controlling the geometric mean is a hyper-parameter: $\lambda_\speaker$ for generation and $\lambda_\listener$ for comprehension. 
In the case of comprehension, the joint probability distribution is:
\begin{align}
    &P_\listener^\joint(\targetindex\vert \imageset, \utterance; \params) = \\ & \hspace{10pt}  \frac{P_\listener(\targetindex\vert\imageset, \utterance;\params)^{\lambda_\listener} P_\speaker(\utterance\vert\imageset, \targetindex;\params)^{1-\lambda_\listener}}{\sum_{t'=1}^N P_\listener(\targetindex'\vert\imageset, \utterance;\params)^{\lambda_\listener} P_\speaker(\utterance\vert\imageset, \targetindex';\params)^{1-\lambda_\listener}} \nonumber\;\;,
\end{align}
where $N$ is the number of targets. 
The joint generation distribution $P_\speaker^\joint(\utterance\vert \imageset, \targetindex; \params)$ is defined in a similar fashion, but with the  $\lambda_\speaker$ hyperparameter. 
Enumerating all possible utterances for the normalization of the joint generation distribution is intractable, so we sample $k$ utterances from $P_\speaker(\utterance\vert \imageset, \targetindex; \params)$ and sum over them to compute the normalization. In the case of comprehension, we can compute the joint distribution exactly because the number of outputs is small (i.e., 10 targets). 
However, if the number of targets was intractably large, the same approximation could also be performed for comprehension.
In practice, we observe the multiplicative generation distribution to skew inference heavily towards short utterances when doing joint inference, and find $\lambda_\speaker=0$ to be the best combination for the joint generation distribution (\autoref{sec:experimental_setup}). Although this eliminates the term $P_\speaker$ from the joint probability, $P_\speaker$ is still influential as the source of samples.

This joint formulation is similar to a rational speech act model~\cite[RSA;][]{goodman2016pragmatic} with a single level of recursion. RSA is a model of pragmatic reasoning, and has been evaluated extensively in reference games~\cite{cohn-gordon-etal-2018-pragmatically, mcdowell-goodman-2019-learning}. We analyze this property for our speaker model in \autoref{sec:results:language}.
Our approximation of the joint speaker distribution is inspired by similar approaches that were applied to RSA~\cite{fried-etal-2018-unified}.

%% file: 50-experimental-setup.tex
\section{Experimental Setup}\label{sec:experimental_setup}

\paragraph{Game Construction} 

We construct  reference game  contexts using the \kilogram dataset~\citep{ji2022abstract} of 1{,}016 abstract tangram shapes. Each context comprises 10 images drawn from this dataset. We use a CLIP model~\citep{radford2021learning} finetuned on \kilogram annotations from~\citet{ji2022abstract}  to ensure visual similarity between images in each context and increase task difficulty. \aautoref{sec:appendix:context} provides further details.

\paragraph{Model and Initialization}

We fine-tune the instruction-tuned \idefics-8B~\citep{laurenccon2024matters}. 
The tasks are delineated via prompting. Training hyperparameters are kept fixed throughout different continual learning rounds and system variants. 
For systems with joint inference, we set $\lambda_L = 0.5$ and $\lambda_S = 0$. 
\aautoref{sec:appendix:model} details prompt design, hyperparameters, and training.
Before the first round of interactions, we initialize the model by fine-tuning \idefics with a small set of 104 successful human-human games. We also add this data to the later rounds of re-training, by assigning all these examples a reward of 1. We use 280 successful human-human games as a validation set for model selection throughout our experiments.

\paragraph{System Variants} 

We refer to our proposed system coupling comprehension and generation with joint inference and data sharing as \variantfull. We compare against three other systems: ablations without data sharing (\variantnods; \autoref{sec:couple:datashare}) or joint inference (\variantnoji; \autoref{sec:couple:jointinference}), and a baseline that uses neither (\variantbaseline). We additionally collect human-human interaction data (\varianthuman) to contextualize performance over time relative to human performance. We also use this human-human data for language analysis.

\paragraph{Deployment} 

We conduct four rounds of deployment, including interactions with human partners and learning. All interactions for each round are collected concurrently in a randomized experiment. 
We collect an equal number of interactions for the speaker and listener roles for each system and round. 
We collect 2{,}000 interactions for each role for each system in the first round, and increase the number by 500 each round, as the marginal benefit of more examples decreases as the data grows.\footnote{Comprehension and generation performance are identical for the \varianthuman system, so we collect half the number of interactions for that system.} 
Because data sharing is not applicable for the first round, the \variantfull and \variantnods, and the \variantnoji and \variantbaseline systems are identical on the first round.
We deploy our systems to interact with human workers on MTurk, at a total cost of \$12,980USD.
\aautoref{sec:appendix:crowdsourcing} provides crowdsourcing details.  

\paragraph{Evaluation} 

At each round, we evaluate comprehension performance from interactions in the listener role using the target selection accuracy. We evaluate a system's generation performance (i.e., as a speaker) as the accuracy of the human interlocutor's target selections. 
For \varianthuman, comprehension and generation performance are identical.

%% file: 60-results.tex
\section{Results and Analysis}\label{sec:results}

We focus on two broad questions: (a) does coupling influence the rate of improvement on task performance (\autoref{sec:results:performance}) and (b) does it lead to quantifiable differences in the generated language over time (\autoref{sec:results:language}).
Overall, we find the answer to both questions is positive, with strong effects. 

\subsection{Performance Analysis}\label{sec:results:performance}

\input{figs/interaction_results}
\input{figs/spatial_reasoning_results}

\autoref{fig:interaction_results} shows model performance over time. 
All systems show dramatic improvement in performance for both comprehension and generation. 
Immediately, we observe significant effect from joint inference, with \variantfull and \variantnods outperforming \variantnoji and \variantbaseline on the first round (53.31\% vs. 42.64\% comprehension, 52.00\% vs. 48.45\% generation).
\variantfull achieves the highest performance at the end of the study, with comprehension improving 53.31$\rightarrow$72.79\% (19.48\% absolute improvement) and generation 52.00$\rightarrow$78.07\% (26.07\% improvement). 
For generation, \variantfull shows the biggest performance delta, even though it starts with already higher performance compared to variants without joint inference. 
With comprehension, \variantnoji (42.64$\rightarrow$66.86\%) shows the biggest delta (24.22\%). 
Coupling dramatically increases learning sample efficiency: \variantfull at the second round already performs better than \variantbaseline at the end of study, even though it trained on less than one third of the data \variantbaseline has seen at the end.

Overall, the gap in performance between \variantfull and \variantbaseline only increases over time. For comprehension, the gap widens 10.67$\rightarrow$14.80\%, but it is much more dramatic for generation with 3.55$\rightarrow$17.10\%. 
Both coupling strategies play a role in this widening  gap in performance, but between the two strategies the relation changes over time. 
Although \variantnods starts with higher performance than \variantnoji, they are essentially equivalent at the end, with \variantnoji showing a trend of outperforming \variantnods. 
This may be because \variantnoji is exposed to more data from the opposing role with data sharing, compensating for the lack of joint inference.\footnote{\autoref{fig:data_sharing_impact} in \aautoref{sec:appendix:data-sharing} depicts data sharing's impact on training set size over time.}

User adaptation is an important potential confounder, potentially explaining any improvements in system performance. 
During the final round, we deploy the initial \variantfull model in a concurrent randomized deployment. 
We observe that human adaptation cannot explain model improvement, seeing  very limited improvement due to adaptation: 0.42\% and 2.56\% for comprehension and generation (cross and dashed curve in \autoref{fig:interaction_results}).

During deployment, a recurring complaint from workers was about the models' inconsistent spatial reasoning, echoing recent evaluations of vision-language models~\citep{kamath-etal-2023-whats, tong2024mass, Tong_2024_CVPR}. 
We identified games where utterances involve a word relating to spatial reasoning.\footnote{\aautoref{sec:appendix:spatial_reasoning} provides the set of words we considered.} 
\autoref{fig:spatial_reasoning_analyses} shows a breakdown of performance trends to games that contain spatial reasoning utterances and games that do not. 
We see a clear difference between the two sets. Although models improve on utterances that contain spatial reasoning, they perform worse on them throughout. 
During the final round, we observe that \variantfull's performance nears that of humans for generation when not using words for spatial reasoning. 

Coupling demonstrates a very strong effect, both on performance and language trends. Balancing the utility of further rounds versus the high cost of each round, we ended the deployment after four rounds. 
\aautoref{sec:appendix:extrapolation} discusses this decision, and provides an extrapolation of performance for one more round, showing a continuation of the observed trends.

\subsection{Language Analysis}\label{sec:results:language}

We study trends in language use over time. 
Throughout this section, except the pragmatic reasoning analysis, we eliminate factors that can complicate the analysis by generating new utterances on the same set of context-target pairs per round for all systems. We randomly sample 2{,}000 context-target pairs from the human-human games for each round, and generate utterances for them with each system using the same inference process as during deployment. 
\autoref{fig:language_analyses} plots the observed trends.\footnote{For all analysis but MAUVE, utterances are lowercased and tokenized with spaCy~\citep{spacy2}.}

We observe a decrease in utterance length for all variants. Humans also show a downward trend in length, likely reflecting the participants becoming experts and therefore more economical in their language. 
This is a known phenomenon in reference games~\citep{krauss1964changes, clark1986referring}, and was also observed in other collaborative scenarios~\citep{effenberger-etal-2021-analysis-language}.
\variantfull and \variantnoji track the human trends best, but generally generate shorter utterances throughout. 

The effective vocabulary of all systems, that is the number of unique words generated for the set of context-target pairs, is also decreasing. 
This has been observed in prior studies for generation systems that are exposed only to their output in continual learning~\cite{kojima2021continual}. 
We expected this effect to be less strong or even reversed once the system is exposed to human utterances, either through data sharing or through joint inference with a comprehension model trained on human utterances. 
The decrease in the vocabulary size is much smaller for the coupled variants, and the smallest for \variantfull, but it remains present. 
We also plot, for each round, how many words a model added to the cumulative set of words it generated until that round (third panel). 
More new words appear for the coupled variants throughout the study.
All systems display a significantly less rich vocabulary compared to humans, leaving an important direction for future work.

We use MAUVE~\citep{pillutla2021mauve}, a reference-less generation evaluation metric, to evaluate the similarity of each model's language to human language. 
For each round and system, we compute the metric between the model- and human-produced utterances for that round. 
We use GPT2-Large as the embedding model~\citep{radford2019language}, similar to \citet{pillutla2021mauve}, and keep the number of clusters fixed at 200. 
We find coupling avoids the drift from human language the \variantbaseline displays. The \variantfull system not only does not stray further from human language, but actually moves closer to it over time. 
Data sharing is particularly critical, but the combination of joint inference further helps to align the model language with human language.

\input{figs/language_analyses}

Finally, we briefly look into whether coupling affects the model's pragmatic reasoning. 
In reference games, the pragmatic information is the images in the context that are not the target. A speaker that employs pragmatic reasoning well will take into account the other images so to help the speaker make the right selection in the specific context they share (i.e., the speaker will refer to properties of the target that specifically distinguish from the other images).
We operationalize this question by measuring the diversity of model descriptions for a specific tangram within different context sets. 
We use the Shape Naming Divergence (SND) metric, introduced by \cite{ji2022abstract} to measure the diversity of human annotations for individual tangrams.\footnote{We describe SND in \aautoref{sec:appendix:snd}.} 
Roughly speaking, high SND means high lexical diversity between the descriptions of a specific image.
For each system and each round, we generate utterances for every context-target pair observed in all human-human games throughout continual learning.\footnote{We cannot compute SND for human participants because of insufficient data per round.} 
We get 10.67 utterances per tangram on average. 
\autoref{fig:language_analyses} (right pane) shows mean SND across all tangrams for each model and round. 
Largely, we observe \variantbaseline's pragmatic ability to collapse over time. Data sharing helps to some degree. 
While we see a decrease in SND over time even when using joint inference, this type of coupling shows much higher SND values throughout, indicating greater diversity of utterances and hence a greater pragmatic effect. 
While this effect tracks the vocabulary size trends in practice, it is independent, even if a diverse vocabulary is a necessary, but insufficient condition. 
That said, this analysis of pragmatic reasoning is rudimentary, and future  in-depth analysis is required to identify the exact qualities of this phenomena and how it correlates with system performance.

\footnotetext[11]{\hypertarget{fn:lang_analysis_confidence}Confidence intervals are computed over $n=10{,}000$ random samples of 2{,}000 context-target pairs for each round.}

%% file: figs/interaction_results.tex
\definecolor{Full}{RGB}{0, 0, 0} %
\definecolor{NoDS}{RGB}{111, 178, 228} %
\definecolor{NoJI}{RGB}{70, 156, 118} %
\definecolor{Baseline}{RGB}{193, 125, 165} %
\definecolor{Human}{RGB}{220, 162, 55} %

\newcommand{\sysvarmarkfull}{*}
\newcommand{\sysvarmarknods}{square}
\newcommand{\sysvarmarknoji}{triangle*}
\newcommand{\sysvarmarkbaseline}{o}
\newcommand{\sysvarmarkhuman}{diamond}

\begin{figure}[t!]
    \centering \footnotesize

\begin{tikzpicture}

\begin{groupplot}[
      group style={
        group size= 2 by 1,
        vertical sep=2,
        horizontal sep=32pt,
        x descriptions at=edge bottom},
      width=0.17\textwidth,
      height=0.24\textwidth,
      ylabel style={yshift=-4.3pt},
      title style={align=center, yshift=6.5pt},
      scale only axis,
      ],
      \nextgroupplot[
        xlabel={Round},
        ylabel style={align=center},
        ylabel={Comprehension Role Accuracy},
        xtick=data,
        xticklabels={1, 2, 3, 4},
        ytick={40, 50, 60, 70, 80, 90},
        extra x ticks={0, 2000, 4500, 7500},
        extra x tick labels={0, 2000, 4500, 7500},
        extra x tick style={xticklabel pos=top},
        title=Cumulative \# Interactions
      ]      
      \coordinate (top) at (rel axis cs:0,1);
      \input{figs/interaction_results_subfigures/comprehension_accuracy}
      \nextgroupplot[
        xlabel={Round},
        ylabel style={align=center},
        ylabel={Generation Role Accuracy},
        xtick=data,
        xticklabels={1, 2, 3, 4},
        ytick={40, 50, 60, 70, 80, 90},
        ymin=35.5,
        extra x ticks={0, 2000, 4500, 7500},
        extra x tick labels={0, 2000, 4500, 7500},
        extra x tick style={xticklabel pos=top},
        title=Cumulative \# Interactions
      ]
      \input{figs/interaction_results_subfigures/generation_accuracy}
    \coordinate (bot) at (rel axis cs:1,0);%
\end{groupplot}

    \path(top|-current bounding box.north) --
      coordinate(legendpos)
      (bot|-current bounding box.north);

    \node[
        matrix of nodes,
        anchor=south,
        draw,
        inner xsep = 0.2em,
        node font=\scriptsize,
        draw
      ]at([yshift=0.5ex, xshift=0ex]legendpos)
      {
        \ref{plots:interaction_full} \variantfull \hspace{0.5em}
        \ref{plots:interaction_noji} \variantnoji \hspace{0.5em}
        \ref{plots:interaction_nods} \variantnods \hspace{0.5em} \\
        \ref{plots:interaction_baseline} \variantbaseline \hspace{0.5em}
        \ref{plots:interaction_human} \varianthuman\\};

\end{tikzpicture}
\caption{Comprehension and generation performance for system variants across four rounds of deployment, with 95\% confidence intervals.\protect\footnotemark{} The top $x$-axis indicates the total number of interactions collected for a role up to the deployment round. Coupling comprehension and generation leads to \variantfull outperforming all ablations throughout.}\label{fig:interaction_results}
\end{figure}
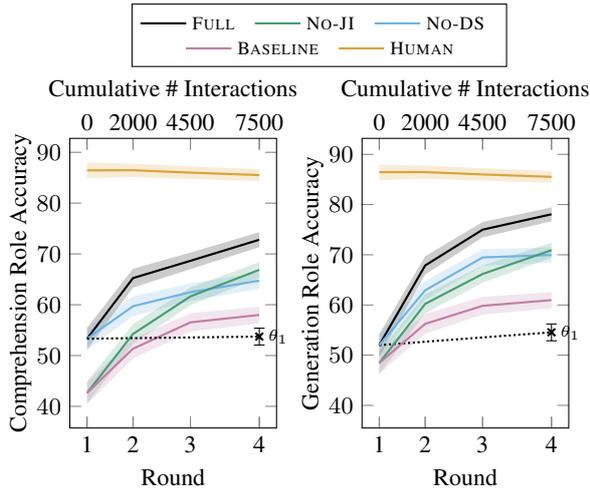

\footnotetext{Confidence intervals are computed using bootstrap sampling, where $n=10{,}000$.}

%% file: figs/interaction_results_subfigures/comprehension_accuracy.tex
\addplot[name path=fullUpperCI, draw=none, mark=none, forget plot] coordinates 
{
(0, 55.47)
(2000, 67.08)
(4500, 70.27)
(7500, 74.28)
};

\addplot[name path=fullLowerCI, draw=none, mark=none, forget plot] coordinates
{
(0, 51.15)
(2000, 63.36)
(4500, 67.0)
(7500, 71.31)
};

\addplot [opacity=0.2, fill=Full, forget plot] fill between[of = fullUpperCI and fullLowerCI];
\addplot [mark=none, Full, thick] coordinates
{
(0, 53.31)
(2000, 65.24)
(4500, 68.64)
(7500, 72.79)
};
\label{plots:interaction_full}

\addplot[name path=noDSUpperCI, draw=none, mark=none, forget plot] coordinates 
{
(0, 55.52)
(2000, 61.62)
(4500, 64.21)
(7500, 66.33)
};

\addplot[name path=noDSLowerCI, draw=none, mark=none, forget plot] coordinates
{
(0, 51.1)
(2000, 57.77)
(4500, 60.71)
(7500, 63.12)
};

\addplot [opacity=0.2, fill=NoDS, forget plot] fill between[of = noDSUpperCI and noDSLowerCI];
\addplot [mark=none, NoDS, thick] coordinates
{
(0, 53.31)
(2000, 59.7)
(4500, 62.47)
(7500, 64.73)
};
\label{plots:interaction_nods}

\addplot[name path=noJIUpperCI, draw=none, mark=none, forget plot] coordinates 
{
(0, 44.84)
(2000, 56.2)
(4500, 63.38)
(7500, 68.4)
};

\addplot[name path=noJILowerCI, draw=none, mark=none, forget plot] coordinates
{
(0, 40.49)
(2000, 52.28)
(4500, 59.91)
(7500, 65.28)
};

\addplot [opacity=0.2, fill=NoJI, forget plot] fill between[of = noJIUpperCI and noJILowerCI];
\addplot [mark=none, NoJI, thick] coordinates
{
(0, 42.64)
(2000, 54.24)
(4500, 61.64)
(7500, 66.86)
};
\label{plots:interaction_noji}

\addplot[name path=baselineUpperCI, draw=none, mark=none, forget plot] coordinates 
{
(0, 44.79)
(2000, 53.3)
(4500, 58.33)
(7500, 59.62)
};

\addplot[name path=baselineLowerCI, draw=none, mark=none, forget plot] coordinates
{
(0, 40.49)
(2000, 49.42)
(4500, 54.69)
(7500, 56.33)
};

\addplot [opacity=0.2, fill=Baseline, forget plot] fill between[of = baselineUpperCI and baselineLowerCI];
\addplot [mark=none, Baseline, thick] coordinates
{
(0, 42.64)
(2000, 51.34)
(4500, 56.53)
(7500, 57.99)
};
\label{plots:interaction_baseline}

\addplot[name path=humanUpperCI, draw=none, mark=none, forget plot] coordinates 
{
(0, 87.99)
(2000, 87.78)
(4500, 87.23)
(7500, 86.68)
};

\addplot[name path=humanLowerCI, draw=none, mark=none, forget plot] coordinates
{
(0, 84.92)
(2000, 85.11)
(4500, 84.73)
(7500, 84.33)
};

\addplot [opacity=0.2, fill=Human, forget plot] fill between[of = humanUpperCI and humanLowerCI];
\addplot [mark=none, Human, thick] coordinates
{
(0, 86.45)
(2000, 86.47)
(4500, 86.0)
(7500, 85.52)
};
\label{plots:interaction_human}

\addplot [
    mark = x, 
    color = Full, 
    thick, 
    nodes near coords = $\theta_1$, 
    every node near coord/.style = {font = \tiny, anchor = 180},
    error bars/.cd,
    y dir = both, %
    y explicit %
] coordinates {
    (7500, 53.73) +-(0, 1.66) %
};

\addplot [color=Full, densely dotted,thick] coordinates {
(0, 53.31)
(7500, 53.73)
};

%% file: figs/interaction_results_subfigures/generation_accuracy.tex
\addplot[name path=fullUpperCI, draw=none, mark=none, forget plot] coordinates 
{
(0, 54.2)
(2000, 69.71)
(4500, 76.53)
(7500, 79.44)
};

\addplot[name path=fullLowerCI, draw=none, mark=none, forget plot] coordinates
{
(0, 49.7)
(2000, 66.07)
(4500, 73.47)
(7500, 76.67)
};

\addplot [opacity=0.2, fill=Full, forget plot] fill between[of = fullUpperCI and fullLowerCI];
\addplot [mark=none, Full, thick] coordinates
{
(0, 52.0)
(2000, 67.87)
(4500, 75.0)
(7500, 78.07)
};

\addplot [
    mark = x, 
    color = Full, 
    thick, 
    nodes near coords = $\theta_1$, 
    every node near coord/.style = {font = \tiny, anchor = 180},
    error bars/.cd,
    y dir = both, %
    y explicit %
] coordinates {
    (7500, 54.56) +=(0, 1.66) -=(0, 1.69) %
};

\addplot [color=Full, densely dotted,thick] coordinates {
(0, 52.0)
(7500, 54.56)
};

\addplot[name path=noDSUpperCI, draw=none, mark=none, forget plot] coordinates 
{
(0, 54.2)
(2000, 64.79)
(4500, 71.1)
(7500, 71.49)
};

\addplot[name path=noDSLowerCI, draw=none, mark=none, forget plot] coordinates
{
(0, 49.8)
(2000, 61.02)
(4500, 67.8)
(7500, 68.46)
};

\addplot [opacity=0.2, fill=NoDS, forget plot] fill between[of = noDSUpperCI and noDSLowerCI];
\addplot [mark=none, NoDS, thick] coordinates
{
(0, 52.0)
(2000, 62.91)
(4500, 69.47)
(7500, 69.97)
};

\addplot[name path=noJIUpperCI, draw=none, mark=none, forget plot] coordinates 
{
(0, 50.65)
(2000, 62.14)
(4500, 67.87)
(7500, 72.42)
};

\addplot[name path=noJILowerCI, draw=none, mark=none, forget plot] coordinates
{
(0, 46.25)
(2000, 58.34)
(4500, 64.43)
(7500, 69.42)
};

\addplot [opacity=0.2, fill=NoJI, forget plot] fill between[of = noJIUpperCI and noJILowerCI];
\addplot [mark=none, NoJI, thick] coordinates
{
(0, 48.45)
(2000, 60.22)
(4500, 66.17)
(7500, 70.91)
};

\addplot[name path=baselineUpperCI, draw=none, mark=none, forget plot] coordinates 
{
(0, 50.7)
(2000, 58.24)
(4500, 61.6)
(7500, 62.57)
};

\addplot[name path=baselineLowerCI, draw=none, mark=none, forget plot] coordinates
{
(0, 46.3)
(2000, 54.28)
(4500, 58.1)
(7500, 59.34)
};

\addplot [opacity=0.2, fill=Baseline, forget plot] fill between[of = baselineUpperCI and baselineLowerCI];
\addplot [mark=none, Baseline, thick] coordinates
{
(0, 48.45)
(2000, 56.24)
(4500, 59.83)
(7500, 60.97)
};

\addplot[name path=humanUpperCI, draw=none, mark=none, forget plot] coordinates 
{
(0, 87.94)
(2000, 87.78)
(4500, 87.23)
(7500, 86.68)
};

\addplot[name path=humanLowerCI, draw=none, mark=none, forget plot] coordinates
{
(0, 84.86)
(2000, 85.11)
(4500, 84.76)
(7500, 84.33)
};

\addplot [opacity=0.2, fill=Human, forget plot] fill between[of = humanUpperCI and humanLowerCI];
\addplot [mark=none, Human, thick] coordinates
{
(0, 86.45)
(2000, 86.47)
(4500, 86.0)
(7500, 85.52)
};

%% file: figs/spatial_reasoning_results.tex
\definecolor{Full}{RGB}{0, 0, 0} %
\definecolor{NoDS}{RGB}{111, 178, 228} %
\definecolor{NoJI}{RGB}{70, 156, 118} %
\definecolor{Baseline}{RGB}{193, 125, 165} %
\definecolor{Human}{RGB}{220, 162, 55} %

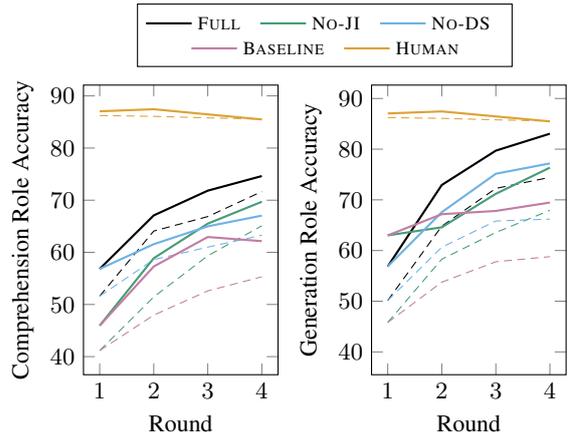
\begin{figure}[t!]
    \centering \footnotesize

\begin{tikzpicture}

\begin{groupplot}[
      group style={
        group size= 2 by 1,
        vertical sep=2,
        horizontal sep=35pt,
        x descriptions at=edge bottom},
      width=0.16\textwidth,
      height=0.24\textwidth,
      scale only axis,
      ],      
      \nextgroupplot[
        xlabel={Round},
        ylabel style={align=center},
        ylabel={Comprehension Role Accuracy},
        xtick=data,
        ytick={40, 50, 60, 70, 80, 90}
      ]
      \coordinate (top) at (rel axis cs:0,1);
      \input{figs/spatial_reasoning_results_subfigures/comprehension_accuracy_with_and_without_spatial_reasoning}
      \nextgroupplot[
        xlabel={Round},
        ylabel style={align=center},
        ylabel={Generation Role Accuracy},
        xtick=data,
        ytick={40, 50, 60, 70, 80, 90},
        ymin=35.5
      ]
      \input{figs/spatial_reasoning_results_subfigures/generation_accuracy_with_and_without_spatial_reasoning}
    \coordinate (bot) at (rel axis cs:1,0);%
\end{groupplot}

    \path(top|-current bounding box.north) --
      coordinate(legendpos)
      (bot|-current bounding box.north);

    \node[
        matrix of nodes,
        anchor=south,
        draw,
        inner xsep = 0.2em,
        node font=\scriptsize,
        draw
      ]at([yshift=0.5ex, xshift=0ex]legendpos)
      {
        \ref{plots:spatial_full} \variantfull \hspace{0.5em}
        \ref{plots:spatial_noji} \variantnoji \hspace{0.5em}
        \ref{plots:spatial_nods} \variantnods \hspace{0.5em} \\
        \ref{plots:spatial_baseline} \variantbaseline \hspace{0.5em}
        \ref{plots:spatial_human} \varianthuman\\};

\end{tikzpicture}
\vspace{-5pt}
\caption{Model comprehension and generation accuracy when the speaker utterance includes ({\protect \tikz
\protect \draw[line width=0.5pt, densely dotted](0,0) -- (0.2,0.2);
}) and does not include ({\protect \tikz
\protect \draw[line width=0.5pt](0,0) -- (0.2,0.2);
}) words for spatial reasoning.
}\label{fig:spatial_reasoning_analyses}

\end{figure}

%% file: figs/spatial_reasoning_results_subfigures/comprehension_accuracy_with_and_without_spatial_reasoning.tex
\addplot [mark=None, Full, densely dashed] coordinates
{
(1, 51.603281133482476)
(2, 64.08083441981746)
(3, 66.85714285714286)
(4, 71.65354330708661)
};

\addplot [mark=None, NoJI, densely dashed] coordinates
{
(1, 41.15942028985507)
(2, 51.445086705202314)
(3, 59.338313767342584)
(4, 65.0904033379694)
};

\addplot [mark=None, NoDS, densely dashed] coordinates
{
(1, 51.603281133482476)
(2, 58.6031746031746)
(3, 60.99629040805512)
(4, 63.28052190121156)
};

\addplot [mark=None, Baseline, densely dashed] coordinates
{
(1, 41.15942028985507)
(2, 47.95982423101067)
(3, 52.60892953200646)
(4, 55.3041018387553)
};

\addplot [mark=None, Human, densely dashed] coordinates
{
(1, 86.23595505617978)
(2, 86.07594936708861)
(3, 85.79023682938617)
(4, 85.54047730463266)
};

\addplot [mark=None, Full, thick] coordinates
{
(1, 56.81470137825421)
(2, 67.08203530633438)
(3, 71.82835820895522)
(4, 74.6268656716418)
};
\label{plots:spatial_full}

\addplot [mark=None, NoJI, thick] coordinates
{
(1, 45.954692556634306)
(2, 58.87353878852285)
(3, 65.50491510277033)
(4, 69.70149253731343)
};
\label{plots:spatial_noji}

\addplot [mark=None, NoDS, thick] coordinates
{
(1, 56.81470137825421)
(2, 61.56351791530945)
(3, 64.98649864986498)
(4, 67.03869047619048)
};
\label{plots:spatial_nods}

\addplot [mark=None, Baseline, thick] coordinates
{
(1, 45.954692556634306)
(2, 57.28476821192053)
(3, 62.940140845070424)
(4, 62.14442013129103)
};
\label{plots:spatial_baseline}

\addplot [mark=None, Human, thick] coordinates
{
(1, 87.04761904761905)
(2, 87.44588744588745)
(3, 86.45720476706393)
(4, 85.48387096774194)
};
\label{plots:spatial_human}

%% file: figs/spatial_reasoning_results_subfigures/generation_accuracy_with_and_without_spatial_reasoning.tex
\addplot [mark=None, Full, densely dashed] coordinates
{
(1, 50.138312586445366)
(2, 64.80411046885035)
(3, 72.25738396624473)
(4, 74.43125618199802)
};

\addplot [mark=None, NoJI, densely dashed] coordinates
{
(1, 45.80378250591016)
(2, 58.29000577700751)
(3, 63.56275303643725)
(4, 67.93286219081273)
};

\addplot [mark=None, NoDS, densely dashed] coordinates
{
(1, 50.138312586445366)
(2, 60.652818991097924)
(3, 65.82832148715146)
(4, 66.2026968247064)
};

\addplot [mark=None, Baseline, densely dashed] coordinates
{
(1, 45.80378250591016)
(2, 53.75552282768778)
(3, 57.78894472361809)
(4, 58.79396984924623)
};

\addplot [mark=None, Human, densely dashed] coordinates
{
(1, 86.23595505617978)
(2, 86.07594936708861)
(3, 85.79023682938617)
(4, 85.54047730463266)
};

\addplot [mark=None, Full, thick] coordinates
{
(1, 56.85920577617328)
(2, 72.92993630573248)
(3, 79.71014492753623)
(4, 83.05084745762711)
};

\addplot [mark=None, NoJI, thick] coordinates
{
(1, 62.98701298701299)
(2, 64.58333333333333)
(3, 71.19140625)
(4, 76.35627530364373)
};

\addplot [mark=None, NoDS, thick] coordinates
{
(1, 56.85920577617328)
(2, 67.56756756756756)
(3, 75.14944491887276)
(4, 77.18567860116569)
};

\addplot [mark=None, Baseline, thick] coordinates
{
(1, 62.98701298701299)
(2, 67.170626349892)
(3, 67.81045751633987)
(4, 69.46778711484593)
};

\addplot [mark=None, Human, thick] coordinates
{
(1, 87.04761904761905)
(2, 87.44588744588745)
(3, 86.45720476706393)
(4, 85.48387096774194)
};

%% file: figs/language_analyses.tex
\definecolor{Full}{RGB}{0, 0, 0} %
\definecolor{NoDS}{RGB}{111, 178, 228} %
\definecolor{NoJI}{RGB}{70, 156, 118} %
\definecolor{Baseline}{RGB}{193, 125, 165} %
\definecolor{Human}{RGB}{220, 162, 55} %

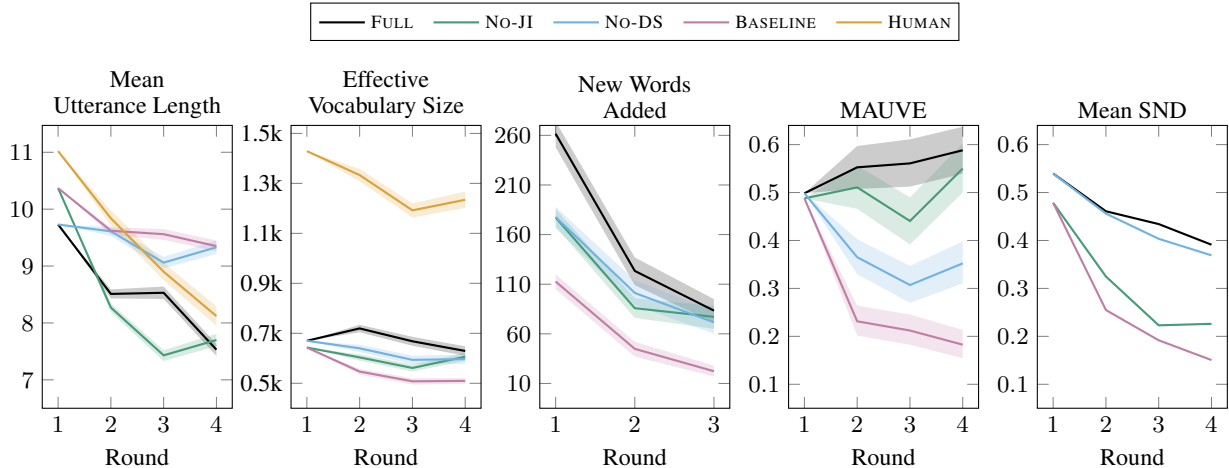
\begin{figure*}[t!]
    \centering \footnotesize

\begin{tikzpicture}

\begin{groupplot}[
      group style={
        group size=5 by 1,
        vertical sep=0,
        horizontal sep=22pt,
        x descriptions at=edge bottom,
        },
      width=2.5cm,
      height=3.75cm,
      scale only axis,
      title style={align=center, yshift=-6.5 pt},
      ],
      \nextgroupplot[
        xlabel={Round},
        ylabel style={align=center},
        xtick=data,
        ytick={7, 8, 9, 10, 11},
        ymin=6.5,
        title=Mean \\ Utterance Length
      ]
      \coordinate (top) at (rel axis cs:0,1);
      \input{figs/language_analysis_subfigures/utterance_length}
      \nextgroupplot[
        xlabel={Round},
        ylabel style={align=center},
        xtick=data,
        ytick={500, 700, 900, 1100, 1300, 1500},
        yticklabels={0.5k, 0.7k, 0.9k, 1.1k, 1.3k, 1.5k},
        ymin=400,
        title=Effective \\ Vocabulary Size
     ]
     \input{figs/language_analysis_subfigures/vocabulary_size}
     \nextgroupplot[
        xlabel={Round},
        ylabel style={align=center},
        xtick=data,
        ytick={10, 60, 110, 160, 210, 260},
        ymin=-15,
        ymax=270,
        title=New Words \\ Added
     ]
     \input{figs/language_analysis_subfigures/new_tokens}
      \nextgroupplot[
        xlabel={Round},
        ylabel style={align=center},
        xtick=data,
        ytick={0.1, 0.2, 0.3, 0.4, 0.5, 0.6},
        ymin=0.05,
        ymax=0.64,
        title=MAUVE
      ]
      \input{figs/language_analysis_subfigures/mauve}
      \nextgroupplot[
        xlabel={Round},
        ylabel style={align=center},
        xtick=data,
        ytick={0.1, 0.2, 0.3, 0.4, 0.5, 0.6},
        ymin=0.05,
        ymax=0.64,
        title=Mean SND
      ]
      \input{figs/language_analysis_subfigures/SND}

    \coordinate (bot) at (rel axis cs:1,0);%
\end{groupplot}

    \path(top|-current bounding box.north) --
      coordinate(legendpos)
      (bot|-current bounding box.north);

    \node[
        matrix of nodes,
        anchor=south,
        draw,
        inner xsep = 0.2em,
        node font=\scriptsize,
        draw
      ]at([yshift=2ex, xshift=0ex]legendpos)
      {
        \ref{plots:language_full} \variantfull \hspace{0.5em}
        \ref{plots:language_noji} \variantnoji \hspace{0.5em}
        \ref{plots:language_nods} \variantnods \hspace{0.5em}
        \ref{plots:language_baseline} \variantbaseline \hspace{0.5em}
        \ref{plots:language_human} \varianthuman\\};

\end{tikzpicture}
\caption{Language analysis plots, with 95\% confidence intervals.\protect\hyperlink{fn:lang_analysis_confidence}{\textsuperscript{11}} Trends in utterance length mirror that of humans when using data sharing (\variantfull and \variantnoji). \variantfull possesses the highest effective vocabulary size and produces the largest number of new words each round.  The \variantfull system additionally shows an increase in MAUVE scores ($\uparrow$) over time and exhibits the highest SND ($\uparrow$) throughout. 
}\label{fig:language_analyses}
\vspace{-5pt}
\end{figure*}

%% file: figs/language_analysis_subfigures/utterance_length.tex
\addplot[name path=fullUpperCI, draw=none, mark=none, forget plot] coordinates 
{
(1, 9.73)
(2, 8.59)
(3, 8.64)
(4, 7.63)
};

\addplot[name path=fullLowerCI, draw=none, mark=none, forget plot] coordinates
{
(1, 9.73)
(2, 8.43)
(3, 8.42)
(4, 7.42)
};

\addplot [opacity=0.2, fill=Full, forget plot] fill between[of = fullUpperCI and fullLowerCI];
\addplot [mark=none, Full, thick] coordinates
{
(1, 9.73)
(2, 8.51)
(3, 8.53)
(4, 7.53)
};
\label{plots:language_full}

\addplot[name path=noJIUpperCI, draw=none, mark=none, forget plot] coordinates 
{
(1, 10.37)
(2, 8.34)
(3, 7.52)
(4, 7.81)
};

\addplot[name path=noJILowerCI, draw=none, mark=none, forget plot] coordinates
{
(1, 10.37)
(2, 8.19)
(3, 7.33)
(4, 7.59)
};

\addplot [opacity=0.2, fill=NoJI, forget plot] fill between[of = noJIUpperCI and noJILowerCI];
\addplot [mark=none, NoJI, thick] coordinates
{
(1, 10.37)
(2, 8.27)
(3, 7.43)
(4, 7.7)
};
\label{plots:language_noji}

\addplot[name path=noDSUpperCI, draw=none, mark=none, forget plot] coordinates 
{
(1, 9.73)
(2, 9.69)
(3, 9.16)
(4, 9.45)
};

\addplot[name path=noDSLowerCI, draw=none, mark=none, forget plot] coordinates
{
(1, 9.73)
(2, 9.53)
(3, 8.96)
(4, 9.21)
};

\addplot [opacity=0.2, fill=NoDS, forget plot] fill between[of = noDSUpperCI and noDSLowerCI];
\addplot [mark=none, NoDS, thick] coordinates
{
(1, 9.73)
(2, 9.61)
(3, 9.06)
(4, 9.33)
};
\label{plots:language_nods}

\addplot[name path=baselineUpperCI, draw=none, mark=none, forget plot] coordinates 
{
(1, 10.37)
(2, 9.68)
(3, 9.65)
(4, 9.45)
};

\addplot[name path=baselineLowerCI, draw=none, mark=none, forget plot] coordinates
{
(1, 10.37)
(2, 9.55)
(3, 9.46)
(4, 9.26)
};

\addplot [opacity=0.2, fill=Baseline, forget plot] fill between[of = baselineUpperCI and baselineLowerCI];
\addplot [mark=none, Baseline, thick] coordinates
{
(1, 10.37)
(2, 9.62)
(3, 9.56)
(4, 9.35)
};
\label{plots:language_baseline}

\addplot[name path=humanUpperCI, draw=none, mark=none, forget plot] coordinates 
{
(1, 11.02)
(2, 9.98)
(3, 9.06)
(4, 8.29)
};

\addplot[name path=humanLowerCI, draw=none, mark=none, forget plot] coordinates
{
(1, 11.02)
(2, 9.71)
(3, 8.75)
(4, 7.95)
};

\addplot [opacity=0.2, fill=Human, forget plot] fill between[of = humanUpperCI and humanLowerCI];
\addplot [mark=none, Human, thick] coordinates
{
(1, 11.02)
(2, 9.84)
(3, 8.9)
(4, 8.12)
};
\label{plots:language_human}

%% file: figs/language_analysis_subfigures/vocabulary_size.tex
\addplot[name path=fullUpperCI, draw=none, mark=none, forget plot] coordinates 
{
(1, 670.0)
(2, 734.0)
(3, 686.0)
(4, 648.0)
};

\addplot[name path=fullLowerCI, draw=none, mark=none, forget plot] coordinates
{
(1, 670.0)
(2, 704.0)
(3, 650.0)
(4, 610.0)
};

\addplot [opacity=0.2, fill=Full, forget plot] fill between[of = fullUpperCI and fullLowerCI];
\addplot [mark=none, Full, thick] coordinates
{
(1, 670.0)
(2, 719.5)
(3, 668.05)
(4, 629.35)
};

\addplot[name path=noJIUpperCI, draw=none, mark=none, forget plot] coordinates 
{
(1, 643.0)
(2, 616.0)
(3, 575.0)
(4, 622.0)
};

\addplot[name path=noJILowerCI, draw=none, mark=none, forget plot] coordinates
{
(1, 643.0)
(2, 591.0)
(3, 547.0)
(4, 590.0)
};

\addplot [opacity=0.2, fill=NoJI, forget plot] fill between[of = noJIUpperCI and noJILowerCI];
\addplot [mark=none, NoJI, thick] coordinates
{
(1, 643.0)
(2, 603.61)
(3, 561.31)
(4, 606.2)
};

\addplot[name path=noDSUpperCI, draw=none, mark=none, forget plot] coordinates 
{
(1, 670.0)
(2, 654.0)
(3, 611.0)
(4, 615.0)
};

\addplot[name path=noDSLowerCI, draw=none, mark=none, forget plot] coordinates
{
(1, 670.0)
(2, 625.0)
(3, 576.0)
(4, 578.0)
};

\addplot [opacity=0.2, fill=NoDS, forget plot] fill between[of = noDSUpperCI and noDSLowerCI];
\addplot [mark=none, NoDS, thick] coordinates
{
(1, 670.0)
(2, 639.87)
(3, 593.73)
(4, 596.68)
};

\addplot[name path=baselineUpperCI, draw=none, mark=none, forget plot] coordinates 
{
(1, 643.0)
(2, 558.0)
(3, 520.0)
(4, 521.0)
};

\addplot[name path=baselineLowerCI, draw=none, mark=none, forget plot] coordinates
{
(1, 643.0)
(2, 535.0)
(3, 495.0)
(4, 498.0)
};

\addplot [opacity=0.2, fill=Baseline, forget plot] fill between[of = baselineUpperCI and baselineLowerCI];
\addplot [mark=none, Baseline, thick] coordinates
{
(1, 643.0)
(2, 546.82)
(3, 507.91)
(4, 509.46)
};

\addplot[name path=humanUpperCI, draw=none, mark=none, forget plot] coordinates 
{
(1, 1429.0)
(2, 1358.0)
(3, 1220.0)
(4, 1267.0)
};

\addplot[name path=humanLowerCI, draw=none, mark=none, forget plot] coordinates
{
(1, 1429.0)
(2, 1308.0)
(3, 1163.0)
(4, 1202.0)
};

\addplot [opacity=0.2, fill=Human, forget plot] fill between[of = humanUpperCI and humanLowerCI];
\addplot [mark=none, Human, thick] coordinates
{
(1, 1429.0)
(2, 1333.26)
(3, 1192.03)
(4, 1234.2)
};

%% file: figs/language_analysis_subfigures/new_tokens.tex
\addplot[name path=fullUpperCI, draw=none, mark=none, forget plot] coordinates 
{
(1, 274.0)
(2, 137.0)
(3, 95.0)
};

\addplot[name path=fullLowerCI, draw=none, mark=none, forget plot] coordinates
{
(1, 248.0)
(2, 109.0)
(3, 71.0)
};

\addplot [opacity=0.2, fill=Full, forget plot] fill between[of = fullUpperCI and fullLowerCI];
\addplot [mark=none, Full, thick] coordinates
{
(1, 261.63)
(2, 123.39)
(3, 83.36)
};

\addplot[name path=noJIUpperCI, draw=none, mark=none, forget plot] coordinates 
{
(1, 186.0)
(2, 96.0)
(3, 88.0)
};

\addplot[name path=noJILowerCI, draw=none, mark=none, forget plot] coordinates
{
(1, 168.0)
(2, 76.0)
(3, 65.0)
};

\addplot [opacity=0.2, fill=NoJI, forget plot] fill between[of = noJIUpperCI and noJILowerCI];
\addplot [mark=none, NoJI, thick] coordinates
{
(1, 177.23)
(2, 85.84)
(3, 77.12)
};

\addplot[name path=noDSUpperCI, draw=none, mark=none, forget plot] coordinates 
{
(1, 188.0)
(2, 114.0)
(3, 83.0)
};

\addplot[name path=noDSLowerCI, draw=none, mark=none, forget plot] coordinates
{
(1, 166.38)
(2, 89.0)
(3, 60.0)
};

\addplot [opacity=0.2, fill=NoDS, forget plot] fill between[of = noDSUpperCI and noDSLowerCI];
\addplot [mark=none, NoDS, thick] coordinates
{
(1, 177.65)
(2, 101.19)
(3, 71.56)
};

\addplot[name path=baselineUpperCI, draw=none, mark=none, forget plot] coordinates 
{
(1, 120.0)
(2, 52.0)
(3, 28.0)
};

\addplot[name path=baselineLowerCI, draw=none, mark=none, forget plot] coordinates
{
(1, 105.0)
(2, 37.0)
(3, 17.0)
};

\addplot [opacity=0.2, fill=Baseline, forget plot] fill between[of = baselineUpperCI and baselineLowerCI];
\addplot [mark=none, Baseline, thick] coordinates
{
(1, 112.64)
(2, 44.84)
(3, 22.31)
};

%% file: figs/language_analysis_subfigures/mauve.tex
\addplot[name path=fullUpperCI, draw=none, mark=none, forget plot] coordinates 
{
(1, 0.4984965591144428)
(2, 0.5971288613298223)
(3, 0.611005118808336)
(4, 0.637248723834267)
};

\addplot[name path=fullLowerCI, draw=none, mark=none, forget plot] coordinates
{
(1, 0.4984965591144428)
(2, 0.5076147456384497)
(3, 0.5122597338245952)
(4, 0.5393087966985025)
};

\addplot [opacity=0.2, fill=Full, forget plot] fill between[of = fullUpperCI and fullLowerCI];
\addplot [mark=none, Full, thick] coordinates
{
(1, 0.4984965591144429)
(2, 0.5525736543595811)
(3, 0.5608109301483083)
(4, 0.5882176391555681)
};

\addplot[name path=noJIUpperCI, draw=none, mark=none, forget plot] coordinates 
{
(1, 0.487734677750259)
(2, 0.5565961814155418)
(3, 0.48988562529848373)
(4, 0.6022069785214915)
};

\addplot[name path=noJILowerCI, draw=none, mark=none, forget plot] coordinates
{
(1, 0.487734677750259)
(2, 0.466049931735672)
(3, 0.3916002443612158)
(4, 0.499852613633357)
};

\addplot [opacity=0.2, fill=NoJI, forget plot] fill between[of = noJIUpperCI and noJILowerCI];
\addplot [mark=none, NoJI, thick] coordinates
{
(1, 0.48773467775025914)
(2, 0.5108228689480279)
(3, 0.44057989339341785)
(4, 0.5500108604265442)
};

\addplot[name path=noDSUpperCI, draw=none, mark=none, forget plot] coordinates 
{
(1, 0.4984965591144428)
(2, 0.4026838297184365)
(3, 0.3469195003525436)
(4, 0.39804674919665084)
};

\addplot[name path=noDSLowerCI, draw=none, mark=none, forget plot] coordinates
{
(1, 0.4984965591144428)
(2, 0.32962747543206944)
(3, 0.26989668981421067)
(4, 0.310435289737464)
};

\addplot [opacity=0.2, fill=NoDS, forget plot] fill between[of = noDSUpperCI and noDSLowerCI];
\addplot [mark=none, NoDS, thick] coordinates
{
(1, 0.4984965591144429)
(2, 0.3650743953759613)
(3, 0.30702839268518156)
(4, 0.3522706214712616)
};

\addplot[name path=baselineUpperCI, draw=none, mark=none, forget plot] coordinates 
{
(1, 0.487734677750259)
(2, 0.26403267626848714)
(3, 0.24539936526439007)
(4, 0.21346535145412254)
};

\addplot[name path=baselineLowerCI, draw=none, mark=none, forget plot] coordinates
{
(1, 0.487734677750259)
(2, 0.20136815057922672)
(3, 0.18241002494453096)
(4, 0.15445954243792112)
};

\addplot [opacity=0.2, fill=Baseline, forget plot] fill between[of = baselineUpperCI and baselineLowerCI];
\addplot [mark=none, Baseline, thick] coordinates
{
(1, 0.48773467775025914)
(2, 0.23137353774913827)
(3, 0.2121743867324888)
(4, 0.1825681765052514)
};

%% file: figs/language_analysis_subfigures/SND.tex
\addplot [mark=None, Full, thick] coordinates
{
(1, 0.53959450094282)
(2, 0.4608924607019649)
(3, 0.4342210591966484)
(4, 0.391136851749862)
};

\addplot [mark=None, NoJI, thick] coordinates
{
(1, 0.47847598857695417)
(2, 0.32480211028556977)
(3, 0.2229891763420593)
(4, 0.22611438380744775)
};

\addplot [mark=None, NoDS, thick] coordinates
{
(1, 0.53959450094282)
(2, 0.4560556561098133)
(3, 0.4034755884377133)
(4, 0.3690022450072706)
};

\addplot [mark=None, Baseline, thick] coordinates
{
(1, 0.47847598857695417)
(2, 0.25485258332369776)
(3, 0.19176906360778298)
(4, 0.15052200692266315)
};

%% file: 70-related-work.tex
\section{Related Work}

Our joint inference strategy (\autoref{sec:couple:jointinference}) is technically based on approximations~\citep{fried-etal-2018-unified, fried2018speaker} of the Rational Speech Acts framework~\citep[RSA;][]{goodman2016pragmatic,yuan2018understanding}, which frames pragmatic reasoning as a recursive process between listener and speaker models. 
RSA has been studied extensively with the focus of developing models that reason pragmatically~\cite[e.g.,][]{monroe-etal-2017-colors,andreas-klein-2016-reasoning}, including through incorporation in learning~\cite{mcdowell-goodman-2019-learning} and inference~\cite{white2020learning}.
We use it for different aims, as one of two strategies to couple comprehension and generation. 
\citet{liu2023computational} studied the incorporation of joint inference for generation learning, which is a component of our study, with a static model listener. 
In contrast, we study learning dynamics  for both comprehension and generation, evaluate data sharing as an additional coupling mechanism, and deploy for continual learning with humans, who constitute non-static partners.

Continually learning from interactions with human users has been studied in the context of instruction generation~\citep{kojima2021continual} and following~\citep{suhr2024continual}, question answering~\citep{gao-etal-2023-continually}, and ad-hoc adaptation~\citep{hawkins-etal-2020-continual}.
In our work, continual learning enables us to study long-term dynamics that arise from coupling comprehension and generation. 
Our continual learning setup is different from the Reinforcement Learning from Human Feedback framework~\citep[RLHF; ][]{ziegler2019fine} in relying on binary signals derived from interactions with users, while RLHF requires external annotators that compare output pairs.

The reference game scenario has been extensively used in cognitive studies as a prototypical, but simple interaction design~\citep{rosenberg1964speakers, krauss1964changes}. 
It has been used to study convention formation at dyadic~\citep{clark1986referring, wilkes1992coordinating} and population-levels~\citep{hawkins2023partners}, and demonstrate computational theories of pragmatic reasoning~\citep{goodman2016pragmatic, cohn-gordon-etal-2019-incremental}, among other behaviors. 
It has also been used to develop computational methods, such as to evaluate contrastive captioning~\citep{vedantam2017context, ou-etal-2023-pragmatic} and abstract reasoning of vision-language models~\citep{ji2022abstract}. The tangram images we use, abstract shapes composed of the same set of seven primitives, likewise have extensive use as stimuli in cognitive science~\citep{clark1986referring, schober1989understanding, horton2002speakers}. They also remain challenging for contemporary models~\citep{ji2022abstract}, making them well suited to demonstrate model improvement.

%% file: 80-discussion.tex
\section{Conclusion}

We study the dynamics of coupling language comprehension and generation at inference- and training-time through a continual learning setting where an agent learns from interactions with humans. 
Coupling has significant impact over time, leading to improved agent performance, sample efficiency, and similarity to human language. 

Our work points to multiple directions for future work, including coupling the processes through the training objective in addition to data at training-time, developing more efficient alternatives to sampling utterances during joint inference for generation, and the study of alternative interaction scenarios, including multi-turn settings where dynamics between comprehension and generation can affect an  interaction throughout its duration. Scaling up our approach and experimental setting to a real-world deployment featuring a wider range of tasks and a broader set of feedback signals, such as natural language feedback, constitutes a particularly important direction.

%% file: 90-limitations-and-ethics.tex
\section*{Limitations}
Our work does not touch on an important factor in deployed systems: the addition of new participants into the system. 
To simplify the crowdsourcing setup, we keep the set of workers fixed during our experiments. 
This does not allow us to observe the effect of new participants joining the population and the impact of the data they create interacting with our agents. 
This is an important direction for future work. 
While our methods are not specifically designed for English, our study is only done in English. 
We restrict the language to English and recruit workers from English-majority locales only. 
This qualifies our findings, both with regard to the language choice and the impact of the culture of the participants. 
These are also important variables for future studies.

Unlike how  RL is usually studied in the research community, our continual learning process involves humans in the loop. This entails restrictions in terms of time and cost. 
We opt for simplicity and choose to train models with a REINFORCE-style policy gradient algorithm~\citep{williams1992simple} and retrain models from scratch on the cumulative set of collected data with each round of continual learning. 
A more extensive (and costly) search over methods might impact results. 
We leave the study of more complex RL algorithms, such as PPO~\citep{schulman2017proximal}, as well as different strategies for incorporating data from previous rounds to future work.

We invested significant effort and resources in running our study for a significant amount of interactions and rounds. While we show consistent trends, it is hard to predict trends at much larger scale (e.g., thousands of rounds or millions of interactions). This is beyond the resource available for this research. That said, even if trends change dramatically with such a long horizon, our approach remains useful for faster learning (i.e., reduce regret) in the early life of the system.

\section*{Ethical Considerations}
Our work studies how the coupling of comprehension and generation affects the dynamics of performance and model language. Through coupling at training time, our model trains on both its own generations alongside generations its human partners produced at interaction time. A naive implementation of this strategy during real-world deployment risks aligning model behavior with the biases of its human interlocutors at best and exposing the model to adversarial actors at worst. Appropriate guardrails or further research for selecting when to apply data sharing should be implemented before deployment is considered to ensure safety.

\section*{Acknowledgements}
This research was supported by ARO W911NF21-1-0106, NSF under grant No. 1750499, a gift
from Open Philanthropy, and a gift from Apple.
We thank the Cornell NLP group and Siddhartha Datta for discussion and comments; Vivian Chen, Gloria Geng, and Anne Wu for feedback on the crowdsourcing pipeline; Vivian Chen and Gloria Geng for sharing  code for data visualization; Ron Eliav and Anya Ji for allowing us to build on top of their reference game interface; and the crowdsourcing workers for their participation.

%% file: 90-appendix.tex
\appendix

\section{Training and Inference Details}\label{sec:appendix:model}
\subsection{Training Hyperparameters}\label{sec:appendix:hyperparams}
We use the instruction-tuned \idefics-8B model~\citep{laurenccon2024matters} for all our experiments, and optimize with AdamW~\citep{loshchilov2018decoupled} with a learning rate of $0.0001$ and a weight decay of $0.1$. Each gradient step is computed over independently sampled minibatches of size 32 for comprehension and generation tasks. 

We use LoRA for finetuning~\citep{hu2021lora}, where $r=16$ and $\alpha=8$. We apply adapters to all feedforward layers in the vision encoder, the modality projection and the perceiver-resampler block, but only to the key, query and value projections of the text decoder. We found applying further adapters for the text decoder to exacerbate overfitting. We load and train models with BF16 precision to reduce memory and compute costs. 

We observe the IPS term for negatively rewarded examples (\autoref{sec:parameter_optimization}) to infrequently attain high values in early epochs during pilot experiments. To increase training stability, we clip the IPS term at 5. During our main experiment, clipping is activated for 2-3\% of negatively rewarded generation examples in the first epoch, with the proportion declining afterwards.

\subsection{Stopping Criterion}\label{sec:appendix:stopping}
Each model is trained for a maximum of 15 epochs.  An epoch is a complete pass over data for the comprehension task. We use patience stopping, ending training when model validation accuracy for the comprehension task does not improve for five epochs. For models with joint inference, we compute validation accuracy with the joint listener model $P_\listener^\joint(\targetindex\vert \imageset, \utterance; \params)$, while for models without joint inference, we compute it with the base listener $P_\listener(\targetindex\vert \imageset, \utterance; \params)$. We exclusively use comprehension accuracy. Pilot experiments showed it correlates well with deployment performance.

\subsection{Hyperparameter Search}
Hyperparameter search is done on the seed initialization data, using comprehension accuracy on the validation set as the metric. We vary learning rates $\{1e-5, 5e-5, 1e-4, 2e-4\}$, weight decay $\{0, 1e-3, 1e-1\}$, LoRA $\alpha$ $\{8, 32\}$ and adapter placements (only key-query-value projections; all forward projections; and all forward projections except for the language decoder, which used key-query-value projections) and prompt designs. The selection of LoRA adapters is the most important hyperparameter, showing a strong impact on overfitting at the data scales we work in. 

The $\lambda_l$ hyperparameter for the joint comprehension distribution is tuned on the seed data with the hyperparameters described on \aautoref{sec:appendix:hyperparams}. We save model checkpoints for each epoch of training and inspect comprehension accuracy values on the validation set for different settings of $\lambda_l$. We find $\lambda_l = 0.5$ consistently perform well. 

We choose  $\lambda_s$ by training models with the joint inference strategy with $\lambda_l = 0.5$ and using the hyperparameters and stopping criterion from \aautoref{sec:appendix:hyperparams} and \aautoref{sec:appendix:stopping}. We sample utterances on the validation set and inspect the re-ranking behavior of the joint generation distribution $P_\speaker^\joint(\utterance\vert \imageset, \targetindex; \params)$ with different $\lambda_s$ values. We observe that the utterance the joint distribution $P_\speaker^\joint(\utterance\vert \imageset, \targetindex; \params)$ ranked as the best was often equivalent to the utterance the base generation distribution $P_\speaker(\utterance\vert \imageset, \targetindex; \params)$ ranked as the most likely. This skew towards the base generation distribution is additionally exacerbated with longer training times. 

To determine $\lambda_s$ in light of this, we probe how accurately the joint generation distribution could rank utterances on the validation set. Specifically, for each context-target pair on the validation set, we measure whether the distribution $P_\speaker^\joint(\utterance\vert \imageset, \targetindex; \params)$ assigned higher probabilities to the ground-truth utterance for that pair than distractor utterances collected for other target images in that context. We vary $\lambda_s$ in $[0, 1]$ with increments of $0.01$ and find that $\lambda_s = 0$ achieved the best accuracy at selecting the ground-truth utterance.

\subsection{Prompt Design}
\input{figs/model_prompts/comprehension_prompt}
\input{figs/model_prompts/generation_prompt}

We use the same model (i.e., same architectures and same parameters) for comprehension and generation and designate which task the model should perform through prompting. 
\autoref{fig:comprehension_prompt} and \autoref{fig:generation_prompt} show the prompts for comprehension and generation.

\subsection{Generation Sampling Details}
We sample utterances autoregressively using a temperature of $\tau=0.7$. 
We sample $k=10$ utterances to generate with the joint inference procedure. To isolate the influence of reranking with the comprehension model, we also sample $k=10$ utterances when not performing joint inference and return the utterance with the highest probability.

\subsection{Computational Resources}
Each model is trained with a single GPU, RTX A6000 or NVIDIA A100. Hyperparameter tuning experiments took 100-200 GPU hours total, while training for the main continual learning experiment took approximately 225 GPU hours. For deployment, on the other hand, 
Models are deployed using RTX A6000 and V100 GPUs, with Ray for inference parallelization~\citep{moritz2018ray}.

\section{Context Construction}\label{sec:appendix:context}
Each reference game round involves a context of size $N=10$ comprising 3 blocks (two of size 3 and one of size 4) of visually similar tangrams. We use a CLIP model~\citep{radford2021learning} finetuned by \citet{ji2022abstract} on annotations from the \kilogram ~dataset to construct these sub-blocks. 
The blocks increase the difficulty of the context, because elements within each block have high visual similarity, making both comprehension and generation more challenging. 

Each similarity block is constructed by randomly sampling a tangram, and sampling the rest of the block members from all other tangrams. The sampling is done using a distribution of normalized similarity scores between the first sampled tangrams and all other tangrams. The similarities are computed using CLIP. 

\section{Experiment Details}
\subsection{Set of Spatial Reasoning Words}\label{sec:appendix:spatial_reasoning}
We curate the set of words relating to spatial reasoning by parsing the set of all human and model-generated utterances using spaCy with the \texttt{en\_core\_web\_sm} pipeline~\citep{spacy2}. We collect the set of all words marked with an ADP (adposition) part-of-speech tag, which predominantly contained terms for spatial reasoning in our task, and manually filtered out the words such as ``like'' that are irrelevant to spatial reasoning. We then added words relating to notions of ``left'' and ``right,'' which were not captured under the ADP tag. 

The full set of words we used was: \nlstring{'from', 'towards', 'thru', 'to', 'through', 'until', 'next', 'above', 'along', 'about', 'out', 'inside', 
    'behind', 'outside', 'forward', 'back', 'around', 'beneath', 'atop', 'up', 'apart', 'near', 'at',
    'below', 'into', 'onto', 'toward', 'past', 'upwards', 'before', 'within', 'against', 'between', 'beside', 'on', 'after', 'by', 'over', 'across', 'down', 'opposite', 'underneath', 'in', 'under', 'left', 'leftward', 'leftwards', 'right', 'rightward', 'rightwards'}.

\subsection{Shape Naming Divergence Metric}\label{sec:appendix:snd}

We analyze pragmatic reasoning using the  Shape Naming Divergence (SND) metric~\citep{ji2022abstract}, which  measures how much the naming of individual tangrams varies across different annotations. We repurpose it  to probe pragmatic reasoning by measuring how much a model's description of a given tangram varies across different contexts. Instead of descriptions from different annotators, we compute SND over descriptions of that tangram in different contexts. This gives insight into the impact of the context (i.e., via pragmatic reasoning) on the description of the individual tangram.

\section{Additional Performance Analyses}
\subsection{Estimating Performance on Future Rounds}\label{sec:appendix:extrapolation}

The decisions of experiment length (i.e., in the number of rounds) requires to balance costs and research utility. 
Our main experiment included four rounds of deployment and learning, which was sufficient to answer our research questions given the dramatic differences between the systems. 
Our data does allow us to estimate performance trends for one more round, without collecting additional data. 
We train models for a fifth round given all the interaction data collected in prior rounds, including the last round of deployment, which provided the final performance numbers. 
We compute offline estimate of comprehension performance using human-model interactions collected on the fourth round by the control system (i.e., the initial \variantfull model), which come from the same distribution of human utterances and are unseen by models in training. 

This estimate indicates the trends we observe are robust, and continue for at least one more round beyond our experiment. Comprehension performance continues to improve for all models (\variantfull: 72.79 $\rightarrow$ 76.79\%; \variantnoji: 66.86 $\rightarrow$ 68.25\%; \variantnods: 64.73 $\rightarrow$ 65.93\%; \variantbaseline: 58.04 $\rightarrow$ 64.25\%). \variantfull still outperforms all other systems by a large margin. Importantly, the performance of \variantfull on the second round remains larger than the improved performance of \variantbaseline (65.24\% > 64.25\%), validating our observation that coupling boosts data efficiency. This indicates that the positive impacts the coupling of comprehension and generation has on performance trends are likely to persist.

\subsection{Impact of Data Sharing on Training Set Size}\label{sec:appendix:data-sharing}
\input{figs/data_sharing}

\autoref{fig:data_sharing_impact} shows how the number of datapoints models train on for comprehension and generation tasks change over time. Coupling with data sharing leads to a strong data augmentation effect for \variantfull and \variantnoji, with the number of datapoints shared from the opposing role increasing as the model performance increases.

\section{Crowdsourcing}\label{sec:appendix:crowdsourcing}

\subsection{Worker Recruitment}
We recruit workers with a minimum HIT (Human Intelligence Task) approval rate of 98\% and at least 1{,}000 approved HITs. We restrict the pool to workers from English-majority locales (United States, Canada, Great Britain, Ireland, Australia, and New Zealand). 
Workers complete a video tutorial and a qualification quiz to qualify for our tasks.\footnote{The quiz may be found in our codebase. The video tutorial is accessible at \url{https://lil-lab.github.io/tangrams-refgame-dev/}.} 
The quiz also includes accepting a consent form.
The consent form details how identifiable information of workers (i.e.,AMT worker IDs) is encrypted, how the collected data would be published, and benefits and risks from participating in the study. We recruit a total of 84 workers. This study was qualified as exempt by Cornell University's Institutional Review Board.

Even with the qualification process, workers that produced low-effort responses or colluded with others entered the worker pool. 
We further estimate the the effectiveness of workers via human-human games. 
We collected a set of $113$ pilot games between humans, where at the end of each HIT, players rated their satisfaction with their partner on a Likert scale from 1--6. We removed workers with an average less than $4$ from the pool and manually reviewed the games of the remaining workers. With this process, we restricted the pool of workers to a set of $50$ experts, $41$ of whom joined our final experiments. We collected our initialization and validation data, and performed our continual learning experiment with this set of experts.

\subsection{Payment Details}
The HIT base pay is \$0.60USD. For each round of reference games played within a HIT, workers receive a bonus of \$0.125USD upon success or \$0.05USD upon failure. 
The estimated hourly pay was \$18.31 USD for games between humans, and \$20.55 USD for games between humans and models at the final round. We set the base pay and bonuses through pilot studies among researchers and tuned the values based on estimates of hourly pay during pilot studies.

\subsection{Game Interface}

\begin{figure*}[t!]
\centering
\includegraphics[width=0.95\textwidth,trim={0, 0, 0, 0},clip]{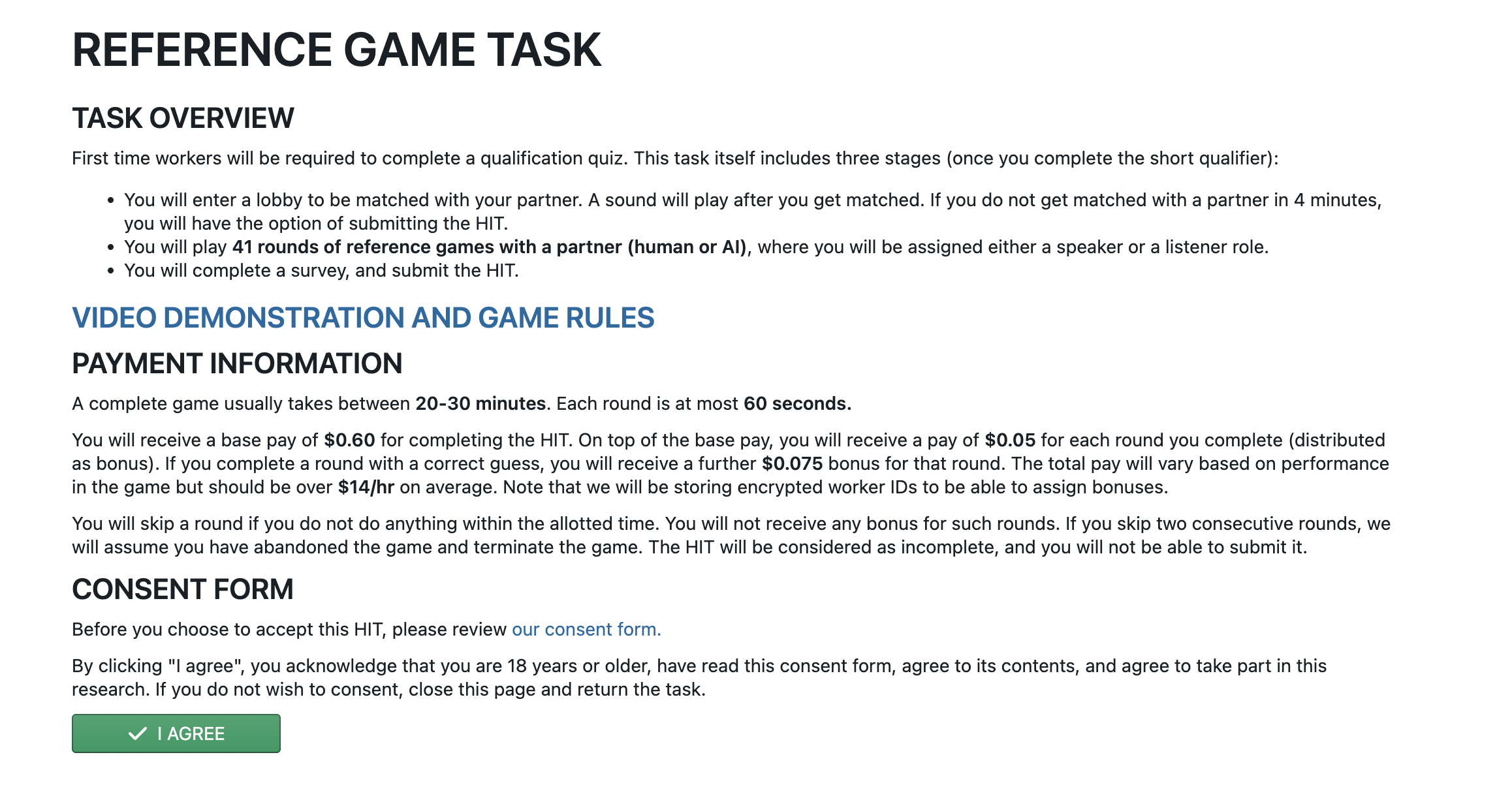}
\includegraphics[width=0.70\textwidth,trim={0, 0, 0, 0},clip]{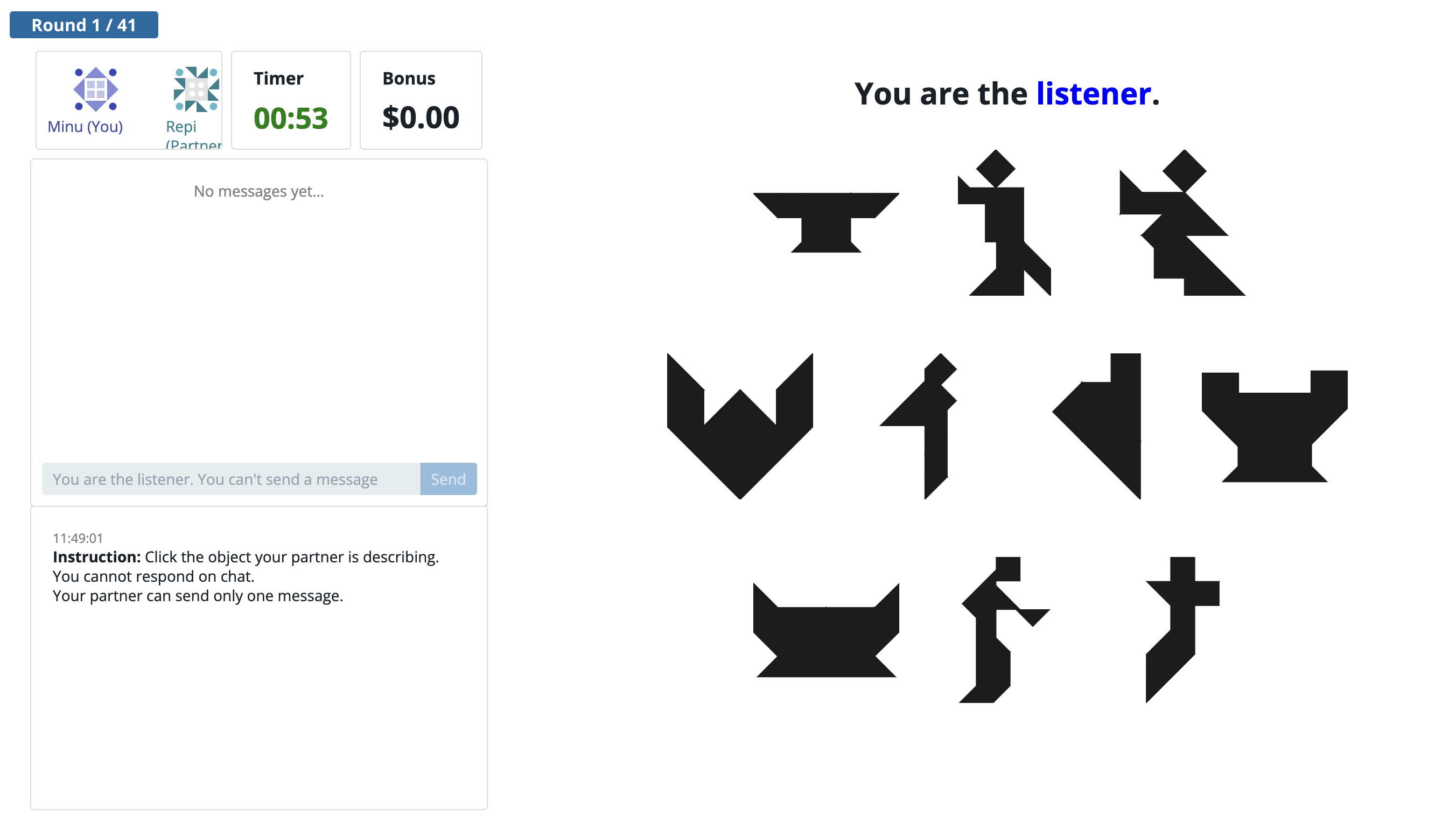}
\includegraphics[width=0.70\textwidth,trim={0, 0, 0, 0},clip]{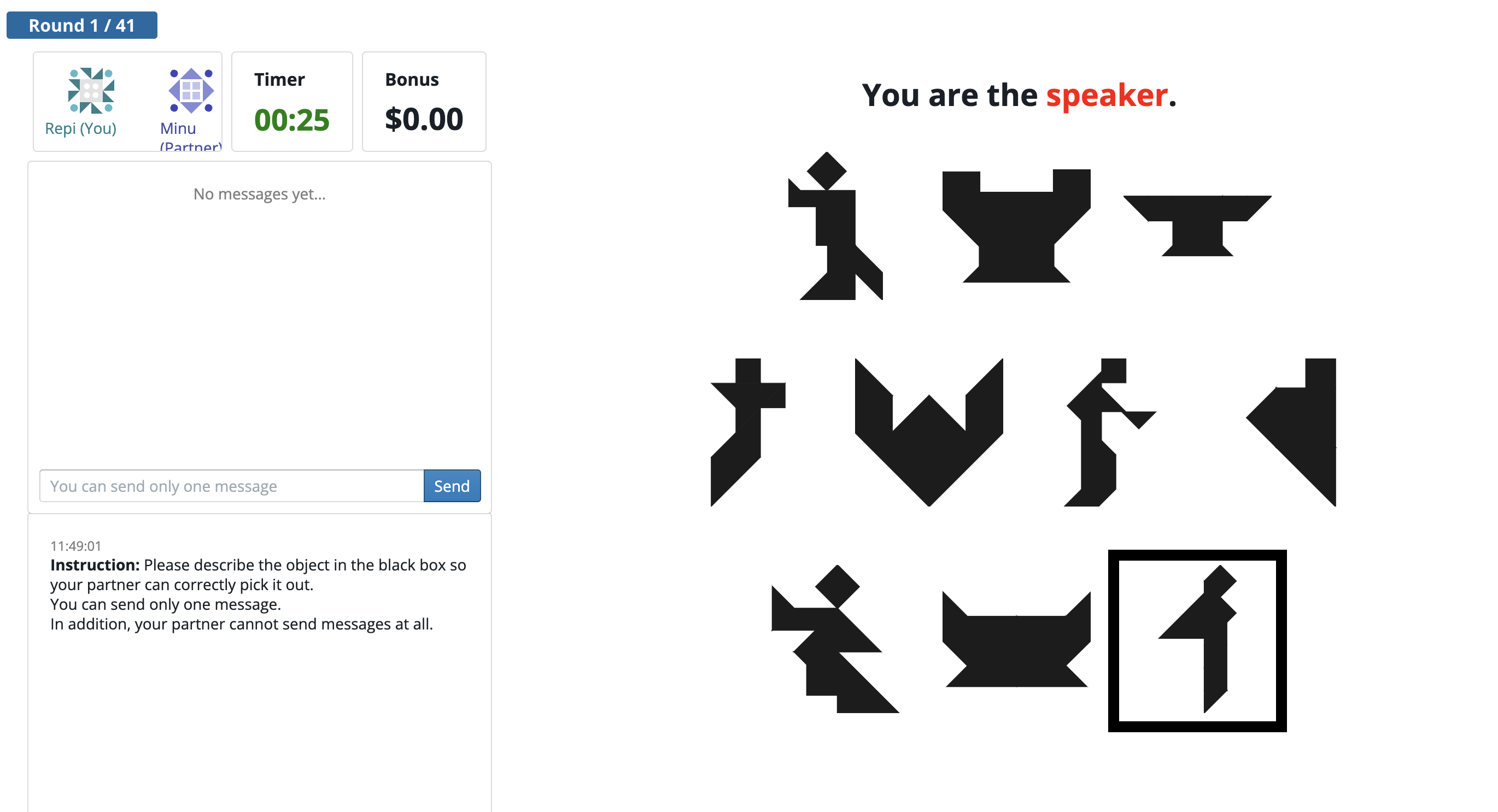}
\caption{Top: the introduction screen shown upon accepting a HIT. Center: worker view in the listener role. Bottom: worker view in the speaker role.}
\label{fig:crowdsourcing_images}
\end{figure*}

The reference game interface is built using the Empirica framework~\citep{almaatouq2021empirica}. It includes a chatbox at the left hand of the screen and the context tangrams at the center. When in the speaker role, the target is indicated to the speaker with a black square. The speaker has 45 seconds to type and send an utterance through the chatbox. 
After the speaker sends a message, the listener is given 15 additional seconds to make a selection. Each round lasts at most 60 seconds. The listener makes a selection by clicking on a tangram image. If successful, the target flashes green for both players. Upon failure, the target tangram flashes red for the speaker and the chosen tangram flashes red for the listener. Workers in the speaker role are not revealed their partners' choice and workers in the listener role are not revealed the target. We do this to mitigate worker adaptation to models and convention formation throughout a HIT. If neither player makes a decision within the given timeframe, the round is considered unsuccessful. The HIT terminates if an individual worker does not take an action for two consecutive rounds. \autoref{fig:crowdsourcing_images} shows the HIT introduction, listener role, and speaker role.

\subsection{Deployment Details}
In each deployment, we give each worker access to an equal number of HITs to uniformly sample from the worker pool. Within a given HIT, a worker plays 40 rounds of reference games, either against a human or model partner. If playing against a model, the worker plays against each system variant an equal number of times and in a random order. During the final round of deployment, we additionally evaluate the initial \variantfull system, and therefore increase the number of rounds per HIT to 50.

Throughout the execution of a HIT, players alternate between roles every 3--4 rounds. In each group of 3--4 rounds, the underlying context is kept fixed, with the targets changing each round. This balances the cognitive load of observing a completely new context while preventing workers from being able to guess targets based on what has not been mentioned yet. If a worker is playing against a model, the system they are playing against is kept fixed within this group of 3--4 rounds. Workers are not revealed whether they are playing against a human or a model.

Each HIT additionally includes an attention check round at a random position. The attention checks are randomly sampled from a set of 100 manually annotated context-target pairs. To ensure simplicity, we sample the targets from the bottom 15th percentile of tangrams in terms of the SND metric (indicating high annotator agreement for tangram naming within the \kilogram dataset) and restrict the remaining tangrams in the context to those with a CLIP cosine similarity less than 0. The rest of the attention check construction follows the process outlined in \aautoref{sec:appendix:context}. 

In practice, we did not  disqualify any workers.
Our main continual learning experiment spanned from May 3rd to May 21st.

\section{Data Details}\label{sec:appendix:data}\subsection{Interactions Per Round}

We collect 2{,}000 interactions for each role for each system in the first round, and increase the number by 500 each round (\autoref{sec:experimental_setup}). For each role and each system, we collect 2{,}000 interactions on round 1, 2{,}500 on round 2, 3{,}000 on round 3, and 3{,}500 on round 4.

\subsection{Data Release}

We release all of the data collected during our experiments alongside the code used to conduct them. This includes the seed training and validation sets of 104 and 280 successful human-human reference games as well as all of the interactions collected during continual learning, comprising 10{,}811 rounds of human-human reference games, and 43{,}442 and 43{,}492 rounds of human-model reference games where the model is in the listener or speaker roles. We do not include rounds where the human partner idled.

During data collection, all worker IDs were encrypted with MD5 hashes. Worker information is further anonymized during the release by mapping each ID hash to a numeric index.

\section{Licenses of Scientific Artifacts Used}
Our chosen model architecture, \idefics-8B~\citep{laurenccon2024matters}, and the Ray library have open licenses (Apache 2.0); the repository for MAUVE~\citep{pillutla2021mauve} has a GNU General Public License; and spaCy~\citep{spacy2} has an MIT license.

%% file: figs/model_prompts/comprehension_prompt.tex
\begin{figure*}[t!]
    \centering
    \fbox{
        \parbox{\dimexpr\textwidth-2\fboxsep-2\fboxrule\relax}{\small
        \textbf{Comprehension Prompt}: \\
\small [User] You will be presented with a sequence of 10 images and a caption describing exactly one of them. Your task is to guess which image the caption describes. Image 0: \textless img0\textgreater, Image 1: \textless img1\textgreater, Image 2: \textless img2\textgreater, Image 3: \textless img3\textgreater, Image 4: \textless img4\textgreater, Image 5: \textless img5\textgreater, Image 6: \textless img6\textgreater, Image 7: \textless img7\textgreater, Image 8: \textless img8\textgreater, Image 9: \textless img9\textgreater. Caption: \textless speaker caption\textgreater. Does this caption describe Image 0, 1, 2, 3, 4, 5, 6, 7, 8 or 9? \\

[Assistant] The caption describes Image \textless target image index\textgreater
        }
    }

    \caption{\idefics comprehension prompt. The target image index is not provided during inference time.}
    \label{fig:comprehension_prompt}
\end{figure*}

%% file: figs/model_prompts/generation_prompt.tex
\begin{figure*}[t!]
    \centering
    \fbox{
        \parbox{\dimexpr\textwidth-2\fboxsep-2\fboxrule\relax}{\small
        \textbf{Generation Prompt}: \\
\small [User] You will be presented with a sequence of 10 images and be assigned a target image. Your task is to produce a caption for your target image such that anyone could guess the image from your description. Image 0: \textless img0\textgreater, Image 1: \textless img1\textgreater, Image 2: \textless img2\textgreater, Image 3: \textless img3\textgreater, Image 4: \textless img4\textgreater, Image 5: \textless img5\textgreater, Image 6: \textless img6\textgreater, Image 7: \textless img7\textgreater, Image 8: \textless img8\textgreater, Image 9: \textless img9\textgreater. Your target is Image \textless image index\textgreater. Produce your caption now.  \\

[Assistant] \textless caption\textgreater
        }
    }

    \caption{\idefics generation prompt. The caption is not provided during inference time.}
    \label{fig:generation_prompt}
\end{figure*}

%% file: figs/data_sharing.tex
\definecolor{Full}{RGB}{0, 0, 0} %
\definecolor{NoDS}{RGB}{111, 178, 228} %
\definecolor{NoJI}{RGB}{70, 156, 118} %
\definecolor{Baseline}{RGB}{193, 125, 165} %
\definecolor{Human}{RGB}{220, 162, 55} %

\begin{figure}[t!]
    \centering \footnotesize

\begin{tikzpicture}

\begin{groupplot}[
      group style={
        group size= 2 by 1,
        vertical sep=2,
        horizontal sep=33pt,
        x descriptions at=edge bottom},
      width=0.16\textwidth,
      height=0.24\textwidth,
      ylabel style={yshift=-4pt},
      scale only axis,
      ],
      \nextgroupplot[
        xlabel={Round},
        ylabel style={align=center},
        ylabel={Comprehension Training Examples},
        xtick=data,
        ytick={0, 2000, 4000, 6000, 8000, 10000, 12000},
        yticklabels={0, 2k, 4k, 6k, 8k, 10k, 12k},
        ymax=13000,
        scaled y ticks=false
      ]      
      \coordinate (top) at (rel axis cs:0,1);
      \input{figs/data_sharing_subfigures/comprehension_datapoints}
      \nextgroupplot[
        xlabel={Round},
        ylabel style={align=center},
        ylabel={Generation Training Examples},
        xtick=data,
        ytick={0, 2000, 4000, 6000, 8000, 10000, 12000},
        yticklabels={0, 2k, 4k, 6k, 8k, 10k, 12k},
        ymax=13000,
        scaled y ticks=false
      ]
      \input{figs/data_sharing_subfigures/generation_datapoints}
    \coordinate (bot) at (rel axis cs:1,0);%
\end{groupplot}

    \path(top|-current bounding box.north) --
      coordinate(legendpos)
      (bot|-current bounding box.north);

    \node[
        matrix of nodes,
        anchor=south,
        draw,
        inner xsep = 0.2em,
        node font=\scriptsize,
        draw
      ]at([yshift=0.5ex, xshift=0ex]legendpos)
      {
        \ref{plots:interaction_full} \variantfull \hspace{0.5em}
        \ref{plots:interaction_noji} \variantnoji \hspace{0.5em} \\
        \ref{plots:interaction_nods} \variantnods \hspace{0.5em} 
        \ref{plots:interaction_baseline} \variantbaseline\\};

\end{tikzpicture}
\caption{Number of training examples for comprehension and generation tasks across four rounds of deployment. The plots account for datapoints converted from the opposing role when data sharing is applied.}\label{fig:data_sharing_impact}
\end{figure}

%% file: figs/data_sharing_subfigures/comprehension_datapoints.tex
\addplot [mark=none, Full, thick] coordinates
{
(1, 104)
(2, 3138)
(3, 7331)
(4, 12578)
};
\label{plots:data_full}

\addplot [mark=none, NoJI, thick] coordinates
{
(1, 104)
(2, 3071)
(3, 7074)
(4, 12052)
};
\label{plots:data_noji}

\addplot [mark=none, NoDS, thick] coordinates
{
(1, 104)
(2, 2098)
(3, 4594)
(4, 7592)
};
\label{plots:data_nods}

\addplot [mark=none, Baseline, thick] coordinates
{
(1, 104)
(2, 2102)
(3, 4601)
(4, 7596)
};
\label{plots:data_baseline}

%% file: figs/data_sharing_subfigures/generation_datapoints.tex
\addplot [mark=none, Full, thick] coordinates
{
(1, 104)
(2, 3167)
(3, 7295)
(4, 12352)
};

\addplot [mark=none, NoJI, thick] coordinates
{
(1, 104)
(2, 2956)
(3, 6810)
(4, 11655)
};

\addplot [mark=none, NoDS, thick] coordinates
{
(1, 104)
(2, 2104)
(3, 4603)
(4, 7603)
};

\addplot [mark=none, Baseline, thick] coordinates
{
(1, 104)
(2, 2104)
(3, 4604)
(4, 7604)
};